\documentclass[runningheads]{llncs}

\usepackage{eccv}

\usepackage{eccvabbrv}

\usepackage{graphicx}
\usepackage{booktabs}

\usepackage[accsupp]{axessibility}  

\usepackage{hyperref}

\usepackage{orcidlink}

\usepackage{amsmath} 
\usepackage{multirow}

\usepackage{amsmath,amsfonts,bm}

\newcommand{\Wtilde}[1]{\stackrel{\sim}{\smash{#1}\rule{0pt}{1.1ex}}}

\def\eqref#1{equation~\ref{#1}}

\def\rvp{{\mathbf{p}}}

\def\rvr{{\mathbf{r}}}

\DeclareMathAlphabet{\mathsfit}{\encodingdefault}{\sfdefault}{m}{sl}
\SetMathAlphabet{\mathsfit}{bold}{\encodingdefault}{\sfdefault}{bx}{n}

\newcommand{\bbR}{\mathbb{R}}

\begin{document}

\title{Ray-Distance Volume Rendering for \\ Neural Scene Reconstruction} 

\titlerunning{Ray-Distance Volume Rendering for Neural Scene Reconstruction}

\author{Ruihong Yin\inst{1}\orcidlink{0000-0001-5989-8976} \and
Yunlu Chen\inst{2}\orcidlink{0009-0001-8958-8490} \and
Sezer Karaoglu\inst{1, 3}\orcidlink{0000-0001-9073-9420} \and
Theo Gevers\inst{1, 3}\orcidlink{0000-0002-1190-5492}}

\authorrunning{R. Yin et al.}

\institute{University of Amsterdam, Amsterdam, The Netherlands \and
Carnegie Mellon University, Pittsburgh, The United States 
 \and
3DUniversum, Amsterdam, The Netherlands \\
\email{\{r.yin, Th.Gevers\}@uva.nl, s.karaoglu@3duniversum.com, yunluche@andrew.cmu.edu}}

\maketitle
\begin{abstract}
  Existing methods in neural scene reconstruction utilize the Signed Distance Function (SDF) to model the density function. However, in indoor scenes, the density computed from the SDF for a sampled point may not consistently reflect its real importance in volume rendering, often due to the influence of neighboring objects. To tackle this issue, our work proposes a novel approach for indoor scene reconstruction, which instead parameterizes the density function with the Signed Ray Distance Function (SRDF). Firstly, the SRDF is predicted by the network and transformed to a ray-conditioned density function for volume rendering. We argue that the ray-specific SRDF only considers the surface along the camera ray, from which the derived density function is more consistent to the real occupancy than that from the SDF. 
  Secondly, although SRDF and SDF represent different aspects of scene geometries, their values should share the same sign {indicating the underlying spatial occupancy}. Therefore, this work introduces a SRDF-SDF consistency loss to constrain the signs of the SRDF and SDF outputs. Thirdly, this work proposes a self-supervised visibility task, introducing the physical visibility geometry to the reconstruction task. The visibility task combines prior from predicted SRDF and SDF as pseudo labels, and contributes to generating more accurate 3D geometry. Our method implemented with different representations has been validated on indoor datasets, achieving improved performance in both reconstruction and view synthesis.
\end{abstract}

\section{Introduction}
\label{sec:intro}

Indoor 3D scene reconstruction involves using multi-view $RGB$ images as input to generate detailed 3D geometry as output. This task is crucial and has widespread applications in virtual/mixed augmentation, robotics, navigation, and so on. Contrary to a single object, indoor environments contain various elements, such as chairs, walls, tables, and cups, which present a challenge for the task of reconstruction. Traditional reconstruction approaches \cite{murez2020atlas, sun2021neuralrecon, bozic2021transformerfusion, yin2023geometry, yao2018mvsnet, sayed2022simplerecon} use 2D and 3D convolutional neural networks (CNNs) \cite{tan2019mnasnet, tang2020searching} to extract features. They can be classified as depth-based methods \cite{yao2018mvsnet, sayed2022simplerecon, duzceker2021deepvideomvs, wang2018mvdepthnet} and volumetric methods \cite{murez2020atlas, sun2021neuralrecon, bozic2021transformerfusion, yin2023geometry, stier2021vortx}. Depth-based techniques estimate depth for every viewpoint, encountering difficulties like scale uncertainties and inconsistencies in depth across different views. The surfaces created through these techniques can be coarse and imprecise. Unlike depth-based approaches, volumetric techniques utilize 3D CNNs to directly generate 3D geometry, leading to smoother surfaces. Nevertheless, these methods use voxels for scene representation, causing GPU memory usage and computational expenses to grow cubically as voxel resolution increases. Such a limitation restricts the use of volumetric approaches for high-quality scene reconstruction. Moreover, volumetric techniques require 3D annotations for network training, significantly increasing the cost of annotations.

Implicit representations \cite{park2019deepsdf, mescheder2019occupancy, sitzmann2020implicit} encode 3D models as functions of coordinates. Utilizing implicit representations and volume rendering, neural radiance fields (NeRF) \cite{mildenhall2021nerf} achieves remarkable performance in novel view synthesis, even without 3D supervision. Some methods \cite{yariv2021volume, wang2021neus} exploit the capabilities of NeRF in reconstruction tasks. These methods typically parameterize the volume density as a learnable transformation of the Signed Distance Function (SDF), leading to impressive reconstruction performance. Compared to classic volumetric methods, neural implicit scene representation is more memory-efficient, marking a new trend in reconstruction tasks.

\begin{figure}[tb]
  \centering
  \includegraphics[width=1.0\linewidth]{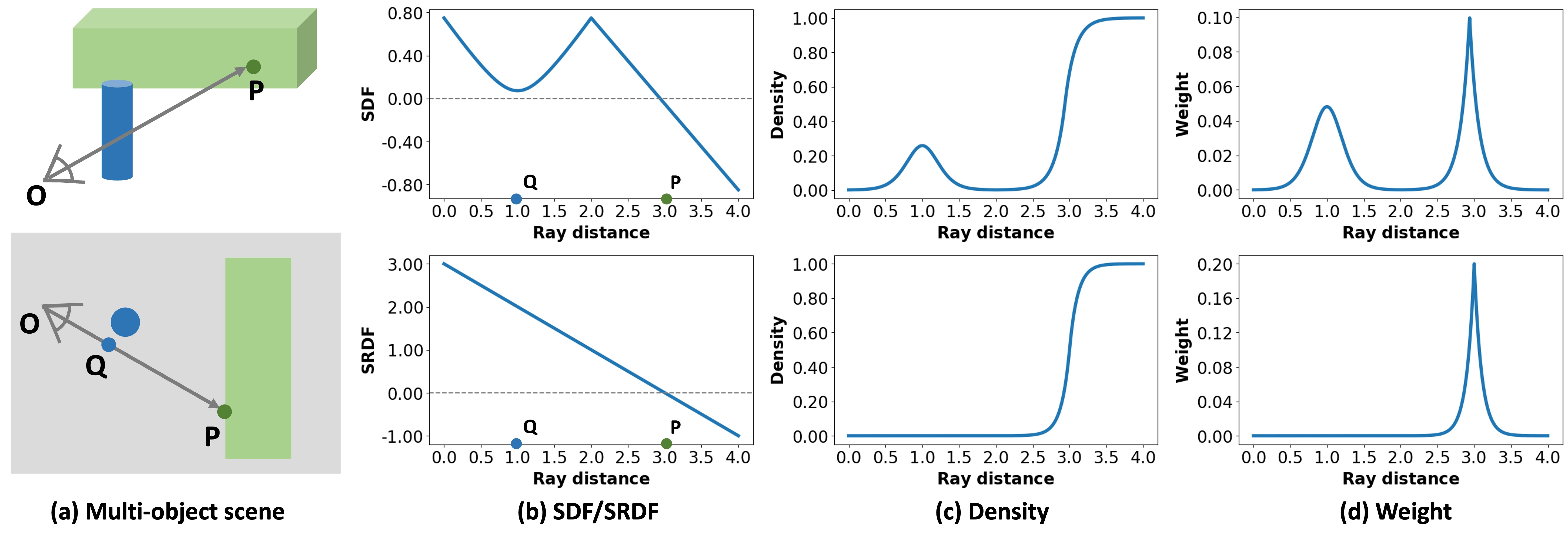}
  \caption{
  \textbf{Analysis of the density function and weight distribution for a toy scene.} 
  (a) Frontal and overhead perspectives of a scene with multiple objects, where a ray originates from the camera center \textbf{O}, intersecting with the green rectangle at the point \textbf{P}. \textbf{Q} is the closest point to the blue cylinder along the ray $\overrightarrow{OP}$. (b) The SDF $d_\Omega$ /SRDF $\Wtilde{d_{\Omega}}$ in \cref{eq:sdf,eq:srdf_sdf} along the ray $\overrightarrow{OP}$. (c) The volume density $\sigma$ generated from the SDF/SRDF in (b) using \cref{eq:density_sdf,eq:density}. (d) The weight distribution $T_i\alpha_i$ generated from the density in (c) via \cref{eq:rendering_color}. Although \textbf{Q} is distant from the surface boundary intersecting with the ray, the density function from the SDF in \textbf{Q} generates a high weight in volume rendering, resulting in noisy rendering and reconstruction. 
  }
  \label{fig:motivation}
\end{figure}

In multi-object indoor scenes, the SDF from sampled points considers surface points across the entire scene. However,  for neural implicit scene reconstruction, along a camera ray, from the camera center to the intersected surface point, the SDF may exhibit fluctuations, leading to several local minimum values. \cref{fig:motivation} provides a toy example to illustrate this. In \cref{fig:motivation}{c}, volume density yielded from the fluctuating SDF displays several local maximum values, contributing to high weights in \cref{fig:motivation}{d}. However, the camera ray $\overrightarrow{OP}$ targets the green rectangle, while the SDF of the near-surface point \textbf{Q} refers to the surface of the blue cylinder. Besides, \textbf{Q} is far from the intersected point \textbf{P}, but still corresponds to high density and high weight. Nevertheless, during volume rendering that maps the 3D output to 2D space, it is evident that along a ray, points closer to the intersected surface point may carry a higher weight as they are more relevant to actual observations. Consequently, in the multi-object scenes, using the SDF to model the density function along the camera ray may disrupt this relationship, introduce noise, and lead to inaccuracies in 2D results or 3D geometries. 

This paper proposes \textbf{RS-Recon}, a novel indoor scene reconstruction method with the ray-specific density function to address the aforementioned issue. Firstly, our method models the density function as a function of the Signed Ray Distance Function (SRDF). Unlike SDF, SRDF measures the shortest distance to the surface along a camera ray, eliminating the influence of surrounding surfaces that are not located on the given ray. Moreover, the density distribution generated from SRDF attains the local maximum only around the intersected surface, aligning more closely to the concept that points nearer to the surface have more importance. To achieve this, our network predicts both SRDF and SDF, where SRDF encompasses the density function, while SDF mainly focuses on describing the 3D surfaces. Secondly, despite SRDF and SDF being defined differently, both of them produce positive values outside the object and negative values within. Therefore, in this paper, a SRDF-SDF consistency loss is devised to ensure they share the same sign. Thirdly, this work introduces a self-supervised visibility task to enhance the prediction of 3D geometry. The visibility task predicts the probability of whether the sampled points are physically visible along the ray in 3D space. The pseudo visibility ground truth is formed by combining the prior of SRDF and SDF from the network itself, without the reliance on multi-view geometry or any additional annotations.

Our contributions can be summarized as follows, \textbf{(1)} This work proposes a novel neural scene reconstruction method that leverages the ray-specific SRDF to model the volume density. Compared to SDF, SRDF better reflects the significance of the real observations. \textbf{(2)} A SRDF-SDF consistency loss is proposed to constrain the alignment between SRDF and SDF, ensuring the same sign inside and outside the object. \textbf{(3)} To enhance the accuracy of 3D geometry, an additional self-supervised visibility task is introduced to predict the visibility probability of sampled points. This task integrates prior information from both the SDF and SRDF predictions as labels. \textbf{(4)} The experiments conducted on both synthetic and real-world indoor datasets illustrate that our method enhances performance in both reconstruction and view synthesis.

\section{Related Works}
\label{sec:relatedwork}

\noindent \textbf{Classic Indoor Scene Reconstruction.} Traditional methods for indoor scene reconstruction fall into two classes, \ie depth-based methods \cite{yao2018mvsnet, sayed2022simplerecon, duzceker2021deepvideomvs, wang2018mvdepthnet, yao2019recurrent, luo2019p, yu2020fast} and volumetric methods \cite{murez2020atlas, sun2021neuralrecon, bozic2021transformerfusion, yin2023geometry, stier2021vortx, ju2023dg, feng2023cvrecon}. Depth-based methods predict pixel-level depth for individual frames, and subsequently depth fusion \cite{curless1996volumetric} is adopted to generate the 3D scene. To achieve favorable outcomes, these methods often construct a plane sweep cost volume \cite{collins1996space, gallup2007real} at the feature or image level, leveraging information from multiple views to complement the current frame, \eg MVSNet \cite{yao2018mvsnet}, DeepVideoMVS \cite{duzceker2021deepvideomvs}, and SimpleRecon \cite{sayed2022simplerecon}. Using the depth map as the intermediate representation, these methods encounter challenges such as depth inconsistency and scale ambiguities. In contrast, volumetric methods utilize 3D CNN to directly regress the 3D geometry, which can generate smooth surfaces and reconstruct unobserved regions. For instance, NeuralRecon \cite{sun2021neuralrecon} designs a learnable TSDF fusion module to integrate features from previous frames and predicts the TSDF for sparse volumes. TransformerFusion \cite{bozic2021transformerfusion}, based on volumetric representation, adopts Transformer to fuse multi-view features and select frames with higher attention. Yin \etal \cite{yin2023geometry} propose to incorporate geometry priors at different levels for the volumetric methods. However, due to the use of 3D CNN in volumetric methods, computational costs escalate significantly with higher resolutions.

\noindent \textbf{Neural Implicit Scene Representation.} With the success of implicit representation and NeRF \cite{mildenhall2021nerf}, neural implicit scene representation has emerged as a hot spot in the past few years. Generally, reconstruction approaches based on neural implicit scene representation first map point positions to a continuous representation of SDF or occupancy by multi-layer perceptrons (MLPs), then transfer SDF or occupancy to volume density, and finally render the 2D results from the 3D outputs. Despite requiring only 2D supervision during training, these methods can reconstruct high-quality surfaces and are more computationally efficient. For example, VolSDF \cite{yariv2021volume} parameterizes the volume density as a Laplace distribution of SDF, while NeuS \cite{wang2021neus} models it as the logistic distribution. Some extensions \cite{yu2022monosdf, guo2022neural, wang2022neuris, wang2022neuralroom, zhu2023nicer} introduce extra priors to achieve superior outcomes. For instance, Manhattan-SDF \cite{guo2022neural} employs planar constraints based on Manhattan-world assumption \cite{coughlan1999manhattan}. NeuRIS \cite{wang2022neuris} proposes to combine normal prior for the reconstruction of texture-less regions, while NeuralRoom \cite{wang2022neuralroom} implements perturbation-residual restriction to reconstruct the flat region better. MonoSDF \cite{yu2022monosdf} exploits both depth and normal cues to improve the reconstruction quality. Although additional priors contribute to the accuracy and completeness of reconstructed surfaces, they employ the SDF to model the volume density, potentially resulting in noisy 2D results in the multi-object scenes. Contrastingly, our method advocates for parameterizing the density function with the ray-specific SRDF, resulting in enhanced reconstruction performance.

Some NeRF-based methods also utilize multi-view geometry to introduce additional regularization. For instance, RegNeRF \cite{niemeyer2022regnerf} regularizes patched-based depth maps from unobserved viewpoints. Geo-NeuS \cite{fu2022geo} and ConsistentNeRF \cite{hu2023consistentnerf} use multi-view photometric consistency. SRDF and observation consistencies are used by \cite{zins2023multi}. GenS \cite{peng2023gens} introduces multi-view and multi-scale feature consistencies for neural surface reconstruction. VIP-NeRF \cite{somraj2023vip} adopts a visibility prior, using the plane sweep volume among multiple views to generate visibility labels. However, they depend on the computation of multi-view geometry, some of which may involve processing multiple frames yielding increased computational demands. In contrast, this paper introduces a self-supervised visibility task that requires minimal computational increases and does not depend on multi-view geometry for ground truth generation.

\begin{figure}[tb]
  \centering
  \includegraphics[width=1.0\linewidth]{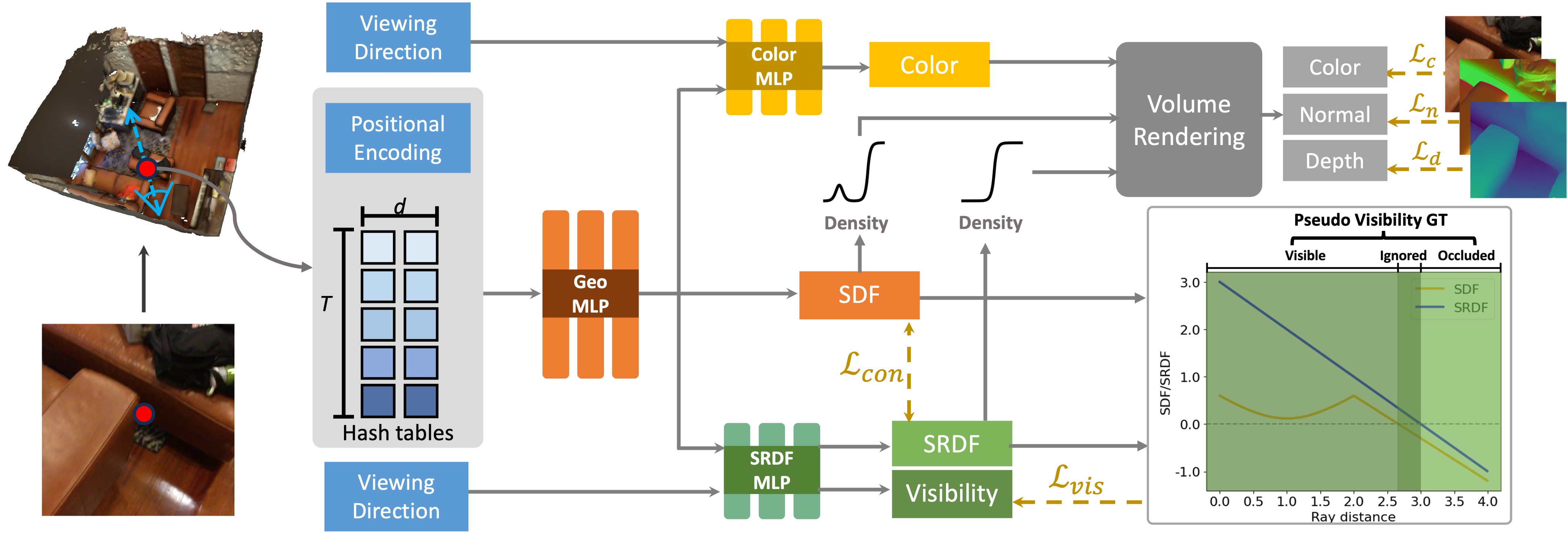}
  \caption{\textbf{Our framework.} A geometry MLP $f_g$ is adopted to generate the SDF and geometry features $\textbf{F}_g$, utilizing the encoded position (and optionally grid features) as input. Then, $\textbf{F}_g$, along with viewing direction, passes through a color MLP $f_c$ to predict the color for each point. Notably, (1) our approach advocates for modeling the density function with the ray-specific SRDF in addition to the SDF. For this purpose, a SRDF MLP $f_s$ is introduced to generate the SRDF. (2) A SRDF-SDF consistency loss $\mathcal{L}_{con}$ is devised to align the signs between the generated SRDF and SDF. (3) To enhance the geometry prediction, a self-supervised visibility task is proposed to integrate geometry priors in both SRDF and SDF predicted by the network and generates the pseudo visibility ground truth. The visibility probability is predicted by the SRDF MLP.
  }
  \label{fig:framework}
\end{figure}

\section{Method}
\label{method}

In this section, we provide the background on NeRF-based reconstruction methods, which typically model the volume density as a transformation of SDF (refer to \cref{sec:preliminary}). To overcome the issue of false local maxima in the density from SDF, our approach covered in \cref{sec:srdf} - \cref{sec:optimization} proposes to model the density function with the ray-specific SRDF in neural scene reconstruction, which achieves more accurate reconstruction. \cref{fig:framework} presents the framework of our approach.

\subsection{Background: Volume Rendering Density as Transformed SDF}
\label{sec:preliminary}

Our goal is to recover the scene geometry given a set of projected 2D views of a 3D scene. To achieve this, NeRF~\cite{mildenhall2021nerf} optimizes a continuous 5D function to represent the scene, which predicts the volume density $\sigma(\rvp)$ and view-dependent color $\textbf{c}(\rvp, \rvr)$ for each sampled point $\rvp \in \bbR$ and ray direction $\rvr \in \mathcal{S}^2$. With $\Omega \subset \bbR^3$ denoting the scene geometry filled with physical presence, the predicted density $\sigma(\rvp)$ approximates the ground truth scene density $\mathbf{1}_{\Omega}(\textbf{p})$, such that $\mathbf{1}_{\Omega}(\textbf{p})=1$ if $\textbf{p} \in \Omega$ and 0 otherwise. 
After that, the classic volume rendering is employed to render the color ${\textbf{C}}(\rvr)$, and $N$ points $\{\textbf{p}_i = \textbf{o} + z_i \mathbf{r}, i\in [1, N]\}$ are sampled along the camera ray $\rvr$, with $z_i$ indicating the depth from the camera center $\textbf{o}$ to the sampled point. The color is accumulated along the ray: 
\begin{equation}
    \begin{aligned}
        {\textbf{C}}(\rvr)=\sum_{i=1}^N T_i\alpha_i \mathbf{c}(\rvp_i, \rvr), 
        T_i=\prod_{j=1}^{i-1}\left(1-\alpha_i\right), 
        \alpha_i=1-\exp \left(-\sigma(\rvp_i) \delta_i\right)
    \end{aligned}
    \label{eq:rendering_color}
\end{equation}
where $\delta_i$ is the interval of neighboring points. 

Despite the great success in novel view synthesis, NeRF has difficulty reconstructing satisfactory actual surfaces from the generated volume density. To address this challenge, recent methods~\cite{yariv2021volume,yu2022monosdf} propose to predict the SDF and then transform it to volume density, which aligns with the geometry bias that the surface points have a higher density than other non-surface points and corresponds to a higher weight during volume rendering. 

For any sampled point $\textbf{p}\in \mathbb{R}^3$, the SDF $d_\Omega(\textbf{p})$ is with an absolute value representing the shortest distance from  $\textbf{p}$ to the surface $\mathcal{M}=\partial\Omega$, while the sign indicates whether the point is outside (positive) or inside (negative) the surface. Formally, 
\begin{equation}
        \text{SDF:}\quad d_{\Omega}(\mathbf{p}) =(-1)^{\mathbf{1}_{\Omega}(\mathbf{p})} \min _{\mathbf{p}^* \in \mathcal{M}}||\mathbf{p}-\mathbf{p}^*||_2,
    \label{eq:sdf}
\end{equation}
where the indicator function $\mathbf{1}_{\Omega}(\textbf{p}) = 1$ if $\textbf{p} \in \Omega$ and 0 otherwise. This indicates that the SDF considers every surface around the point. To apply SDF to volume rendering, Yariv 
\etal ~\cite{yariv2021volume} suggest to derive the volume density $\sigma^{\text{SDF}}$ from the signed distance as follows:
\begin{equation}
    \begin{aligned}
        \sigma^{\text{SDF}}(\textbf{p})=\alpha {\rm{\Psi}}_\beta\left(-{d}_{\Omega}(\textbf{p})\right), 
        {\rm{\Psi}}_\beta(s)= \begin{cases}\frac{1}{2} \exp \left(\frac{s}{\beta}\right) & \text { if } s \leq 0 \\ 1-\frac{1}{2} \exp \left(-\frac{s}{\beta}\right) & \text { if } s>0\end{cases}
    \end{aligned}
    \label{eq:density_sdf}
\end{equation}
in which $\alpha, \beta$ are learnable parameters. ${\rm{\Psi}}_\beta$ is the Cumulative Distribution Function (CDF) of
the Laplace distribution with zero mean and $\beta$ scale. Then, $\sigma^{\text{SDF}}(\textbf{p})$ is applied to \cref{eq:rendering_color} and as the density function $\sigma$. 

\subsection{Accurate Scene Volume Density from Ray-Specific SRDF}
\label{sec:srdf}
We illustrate the issues with volume rendering SDF by a toy example in \cref{fig:motivation}. Indoor scenes typically include multiple objects. Along the camera ray $\mathbf{r}$, the SDF may be influenced by surfaces not intersecting the ray, resulting in several ambiguous local maxima of false high volume density and weight. This implies that the 2D observations may have a strong correlation to these distant points, contradicting the fact that the points closer to the surface along the camera ray may carry more significance. 

To this end, we propose volume rendering scene density function from SRDF, the distance field local to the ray direction. In contrast to the SDF in \cref{eq:sdf}, the SRDF $\Wtilde{d_{\Omega}}(\mathbf{p}, \mathbf{r})$ \cite{zins2023multi, ren2023volrecon} computes the shortest distance from the point to the surface along the sampled ray $\mathbf{r}$, \ie the SRDF is ray-dependent. Formally, the SRDF $\Wtilde{d_{\Omega}}(\mathbf{p},\mathbf{r})$ can be represented by 
\begin{equation}
    \begin{aligned}
        \text{SRDF: }\quad \Wtilde{d_{\Omega}}(\mathbf{p}, \mathbf{r}) =(-1)^{\mathbf{1}_{\Omega}(\mathbf{p})} \min _{(\mathbf{p} + \rho \mathbf{r}) \in \mathcal{M} ;  \rho \in \mathbb{R}} |\rho|,
    \end{aligned}
    \label{eq:srdf_sdf}
\end{equation}
where $\mathbf{1}_{\Omega}(\textbf{p}) = 1$ if $\textbf{p} \in \Omega$ and 0 otherwise. In the following sections, $d_{\Omega}$ from \cref{eq:sdf} and $\Wtilde{d_{\Omega}}$ are used to denote SDF and SRDF for brevity.

In \cref{fig:motivation}, it can be seen that the weight computed from SRDF in the volume rendering only reaches a local peak around the surface point, aligning more consistently with the nature of the 2D observations. Therefore, this paper proposes to employ the ray-specific SRDF to yield the volume density instead of the SDF. This work predicts both SDF and SRDF, where SDF is mainly used to model the surface. As depicted in \cref{fig:framework}, for a sampled point $\textbf{p}$, a geometry MLP $f_g$ is applied on the input (including the encoded position or features) to predict SDF $d_{\Omega}$ and geometry feature $\textbf{F}_g$. By definition, SRDF is view-dependent. Hence, the viewing direction $\mathbf{r}$, geometry feature $\textbf{F}_g$, and point position $\textbf{p}$ are concatenated and then passed through a SRDF MLP $f_s$ to yield the SRDF. On the other hand, the viewing direction $\mathbf{r}$, geometry feature $\textbf{F}_g$, point position $\textbf{p}$, and 3D unit normal $\textbf{n} = \nabla{d_{\Omega}}$ are used as input to a color MLP $f_c$ to output color $\textbf{c}$. 

For volume rendering, our method adopts the predicted SRDF $\Wtilde{d_{\Omega}}$ to derive the volume density $\sigma$. This is defined by:
\begin{equation}
    \begin{aligned}
        \sigma^\text{SRDF}(\textbf{p},\mathbf{r})=\alpha {\rm{\Psi}}_\beta\left(-\Wtilde{d_{\Omega}}(\textbf{p},\mathbf{r})\right).
    \end{aligned}
    \label{eq:density}
\end{equation}
Then, $\sigma^\text{SRDF}(\textbf{p},\mathbf{r})$ is applied to the volume rendering in \cref{eq:rendering_color} as a ray-conditioned density function to obtain the color value ${\textbf{C}}^\text{SRDF}(\rvr)$.  Besides, during training, we also render color ${\textbf{C}}^\text{SDF}(\rvr)$ with density functions from SDF to obtain gradient signals to optimize the prediction of SDF, in which the volume rendering processes use the same rays and points as those for ${\textbf{C}}^\text{SRDF}(\rvr)$.

For the scene geometry, after the network is trained with NeRF optimization targets, the geometric surface $\mathcal{M} = \partial \Omega$ is extracted as watertight meshes by the marching cubes algorithm~\cite{lorensen1998marching,park2019deepsdf} from the SDF results $d_{\Omega}$, which only depends on the query location $\rvp$.

\subsection{SRDF-SDF Consistency Loss}
\label{sec:consistency}
Following the definitions in \cref{eq:srdf_sdf}, despite SRDF and SDF being defined differently, they share the same sign, which indicates the spatial occupancy — positive outside the surface and negative inside the surface.  However, in this paper, where SDF and SRDF are predicted from separate branches, sign consistency is not guaranteed. To deal with this issue, a SRDF-SDF consistency loss is proposed to enforce this constraint. 

Although the sign function effectively computes the output's sign, it cannot be differentiated during back-propagation. To a certain degree, the values derived from the sigmoid function approximate those calculated by the sign function. Thus, in our SRDF-SDF consistency loss, the sigmoid function is first employed to generate the sign of SRDF and SDF as the approximated sign function $\varsigma(\cdot)$, after which a $\ell_2$ loss is adopted to quantify the sign difference, as defined by:
\begin{equation}
    \begin{aligned}
        \mathcal{L}_{con} &= \frac{1}{N_r}\sum_{\textbf{p}, \mathbf{r}} M_{con}||\varsigma(\Wtilde{d_{\Omega}})-\varsigma({d}_{\Omega})||_2  \\
        \varsigma(d) &= \texttt{Sigmoid}(k \cdot d) , 
        M_{con} = [\Wtilde{d_{\Omega}} \cdot {d}_{\Omega} < 0]
    \end{aligned}
    \label{eq:consistency_loss}
\end{equation}
in which $k$ is a hyperparameter, controlling the slope of the sigmoid function. The indicator function $[\Wtilde{d_{\Omega}} \cdot {d}_{\Omega} < 0]$ is leveraged to identify whether the generated SDF and SRDF have opposite signs. Therefore, the consistency loss imposes penalties when there is a difference between the signs of SRDF and SDF. ${N_r}$ represents the number of sampled points in a minibatch that satisfy $[\Wtilde{d_{\Omega}} \cdot {d}_{\Omega} < 0]$.

Taking the derivative of $\mathcal{L}_{con}$ with respect to SDF ${d}_{\Omega}$ (as an example), is computed as follows: 
\begin{equation}
    \begin{aligned}
        \frac{\partial {\mathcal{L}_{con}}}{\partial {d}_{\Omega}} = -\frac{2}{N_r}M_{con}||\varsigma(\Wtilde{d_{\Omega}})-\varsigma({d}_{\Omega})||_1  \cdot
        \frac{\partial {\varsigma({d}_{\Omega})}}{\partial {d}_{\Omega}}
    \end{aligned}
    \label{eq:derivative_consistency}
\end{equation}
The derivative in \cref{eq:derivative_consistency} shows that our consistency loss has two advantages: (1) Unlike the sign function, which assigns the same penalty for outputs with different signs of SRDF and SDF, our consistency loss can adjust the penalty according to the extent of the sign discrepancy between SRDF and SDF, referring to $||\varsigma(\Wtilde{d_{\Omega}})-\varsigma({d}_{\Omega})||_1$. (2) The derivative of the sigmoid function $\frac{\partial {\varsigma({d}_{\Omega})}}{\partial {d}_{\Omega}}$ peaks at zero and decreases from 0 to $+\infty/-\infty$. Consequently, predictions in proximity to zero (indicating the surface) may exhibit higher absolute gradient values, providing effective supervision for points near the surface.

\subsection{Self-supervised Visibility Task}
\label{sec:visibility}
Theoretically, with visibility attributes of 3D points, surface location becomes straightforward. Therefore, to enhance 3D geometry prediction, we propose a self-supervised visibility task in this paper. Along the trajectory of a camera ray, points sampled prior to reaching the first surface point are categorized as visible, whereas points sampled beyond the first surface point are classified as occluded. To find the first surface point along the ray, using SRDF as an example, the multiplication of SRDF between adjacent sampled points is computed. When $\Wtilde{d_{\Omega}}^i \cdot \Wtilde{d_{\Omega}}^{i+1} \le 0$ and ${\Wtilde{d_{\Omega}}^m \cdot \Wtilde{d_{\Omega}}^{m+1}>0, m \in [1, i-1]}$, it indicates that the first surface is within the interval $[\textbf{o} + z_i \mathbf{r}, \textbf{o} + z_{i+1} \mathbf{r}]$. Consequently, the points between $[\textbf{p}_1, \textbf{p}_{i}]$ are considered physically visible while points between $[\textbf{p}_{i+1}, \textbf{p}_{N}]$ are regarded as occluded. However, the SRDF or SDF predicted by the network may be inaccurate at some points, introducing potential noise in the visibility task. To improve the accuracy of the visibility labeling, both SDF and SRDF are utilized in determining visibility. Points are labeled as visible or occluded only when there is agreement between SRDF and SDF on their classification. If the classifications from SRDF and SDF diverge, suggesting difficulties in accurately categorizing these points for the network, such points are omitted from the visibility task. The method for assigning visibility labels is defined as follows
\begin{equation}
    \begin{aligned}
        V_{gt}= \begin{cases} 1, & \text { if } (V^\text{SRDF} = 1) \& (V^\text{SDF} = 1)  \\0, & \text { if } (V^\text{SRDF} = 0) \& (V^\text{SDF} = 0) \end{cases}
    \end{aligned}
    \label{eq:visibility_gt}
\end{equation}
where $V^\text{SRDF}/V^\text{SDF}$ are the visibility labels determined based on the prediction of SRDF/SDF. $1/0$ means label for visible/occluded points. The formation of the visibility label can be inferred as self-supervised, thus eliminating the necessity for multi-view geometry and additional annotations. Additionally, it incorporates information from both SRDF and SDF.

From the analysis above, similar to SRDF, the visibility task also relies on the view direction. Therefore, a visibility probability $V_{pred}$ is predicted from the SRDF branch. It is a binary classification task, \ie visible or occluded, so the binary cross-entropy loss is employed as visibility loss $\mathcal{L}_{vis}$ during optimization.

The visibility task can discern the point's visibility along the camera ray, thereby influencing the learning of SRDF. Moreover, the optimization process integrates knowledge of ray-related visibility across the entire space, further impacting the learning of SDF. Notably, unlike 2D supervision, the visibility task provides priors in 3D space, imparting a more direct influence on 3D geometry.

\begin{table}[tb]
\small
  \caption{Evaluation of 3D reconstruction meshes on ScanNet. $^\dagger$ indicates our reproduced results. The best results are \textbf{bold}. $\downarrow$ denotes that a lower value is preferable while $\uparrow$ means higher value corresponds to better performance.
  }
  \label{tab:scannet}
  \tabcolsep=5pt
  \centering
 
  \begin{tabular}{cccccc}
    \toprule
    Method  & Acc $\downarrow$ & Comp $\downarrow$ &Prec $\uparrow$ &Recall $\uparrow$ &F-score $\uparrow$   \\
    \hline
    
    COLMAP \cite{schonberger2016pixelwise}&  0.047 &0.235 &0.711 &0.441 &0.537  \\
    UNISURF \cite{oechsle2021unisurf} &  0.554 &0.164 &0.212 &0.362 &0.267 \\
    NeuS \cite{wang2021neus}&  0.179 &0.208 &0.313 &0.275 &0.291 \\
    VolSDF \cite{yariv2021volume}&  0.414 &0.120 &0.321 &0.394 &0.346 \\
    Manhattan-SDF \cite{guo2022neural}&  0.072 &0.068 &0.621 &0.586 &0.602 \\
    S3P \cite{ye2023self}&  0.055 &0.059 &0.709 &0.660 &0.683 \\
    NeuRIS \cite{wang2022neuris}&  0.050 &0.049 &0.717 &0.669 &0.692 \\
    MonoSDF\_Grid \cite{yu2022monosdf} &  0.072 &0.057 &0.660 &0.601 &0.626 \\
    MonoSDF\_MLP \cite{yu2022monosdf} &  \textbf{0.035} &0.048 &0.799 &0.681 &0.733 \\
    HelixSurf \cite{liang2023helixsurf} &  {0.038} &0.044 &0.786 &0.727 &0.755 \\
    Occ\_SDF\_Hybrid$^\dagger$ \cite{lyu2023learning} &  {0.040} &0.041 &0.783 &0.748 &0.765 \\
    \hline
    Ours\_Grid  &  {0.074} &{0.049} &{0.670} &{0.703} &{0.683} \\
    Ours\_MLP  &  {0.040} &\textbf{0.040} &\textbf{0.809} &\textbf{0.779} &\textbf{0.794} \\
    \bottomrule
  \end{tabular}
\end{table}

\subsection{Optimization}
\label{sec:optimization}
\noindent \textbf{Loss Functions.} During training, the overall loss consists of RGB loss $\mathcal{L}_c$, normal loss $\mathcal{L}_n$, depth loss $\mathcal{L}_d$, Eikonal loss $\mathcal{L}_e$, smooth loss $\mathcal{L}_s$, SRDF-SDF consistency loss $\mathcal{L}_{con}$, and visibility loss $\mathcal{L}_{vis}$, 
\begin{equation}
    \begin{aligned}
        \mathcal{L} = \mathcal{L}_c + \lambda_n \mathcal{L}_n + \lambda_d \mathcal{L}_d + \lambda_e \mathcal{L}_e + \lambda_s \mathcal{L}_s + \lambda_{con} \mathcal{L}_{con} + \lambda_{vis} \mathcal{L}_{vis}
    \end{aligned}
    \label{eq:total_loss}
\end{equation}
in which $\lambda_{(\cdot)}$ are the loss weight. Following MonoSDF \cite{yu2022monosdf}, the 2.5D depth and normal maps are volume rendered using the density function $\sigma^\text{SRDF}$ to exploit geometric cues and support training. More details on the computation of loss functions are provided in the supplementary material. 

\noindent \textbf{Implementation Details.} Theoretically, our method can be applied to replace SDF-based volume rendering in indoor scene reconstruction. To measure our method, VolSDF-based reconstruction method \emph{MonoSDF} is considered as our baseline in the following experiments. Furthermore, in the supplementary material, we also verify our method on NeuS-based reconstruction method \emph{NeuRIS}. We used different spatial encoding techniques as input, including hash features \cite{muller2022instant} and Fourier features \cite{mildenhall2021nerf}, referred to as \emph{Grid} and \emph{MLP} respectively in the following experiments. More details are given in the supplementary material.

\section{Experiments}
\label{sec:exp}

\subsection{Experimental Setup}
\label{sec:exp setup}
\noindent \textbf{Datasets.}
Our proposed method is evaluated on three indoor datasets, comprising: (1) four scenes from the real-world dataset ScanNet \cite{dai2017scannet}, (2) eight scenes from the synthetic dataset Replica \cite{straub2019replica}, and (3) four scenes from the advanced set of the real-world large-scale dataset Tanks and Temples \cite{knapitsch2017tanks}. 

\noindent \textbf{Metrics.} For ScanNet, the assessment of the results involves five reconstruction metrics: accuracy (acc), completeness (comp), precision (prec), recall, and F-score with a threshold of 5cm. For Replica, in alignment with MonoSDF, metrics including normal consistency (normal c.), chamfer L1 distance (Chamfer-L1), and F-score with a threshold of 5cm are utilized. For Tanks and Temples, as evaluated by the official server, results are reported based on the F-score with a 1cm threshold. Additionally, this study assesses the peak signal-to-noise ratio (PSNR) of the rendered 2D $RGBs$ to showcase the effectiveness of our approach. Detailed computations for all metrics are given in the supplementary material.

\noindent \textbf{Baselines.} Except for the baseline MonoSDF, our method is also compared to (1) the classic MVS method: COLMAP \cite{schonberger2016pixelwise}, (2) neural implicit reconstruction methods: UNISURF \cite{oechsle2021unisurf}, NeuS \cite{wang2021neus}, VolSDF \cite{yariv2021volume}, Manhattan-SDF \cite{guo2022neural}, NeuRIS \cite{wang2022neuris}, S3P \cite{ye2023self}, HelixSurf \cite{liang2023helixsurf}, and Occ\_SDF\_Hybrid \cite{lyu2023learning}.

\subsection{Evaluation Results}
\label{sec: exp sota}

\begin{figure}[tb]
  \centering
  \includegraphics[width=0.94\linewidth]{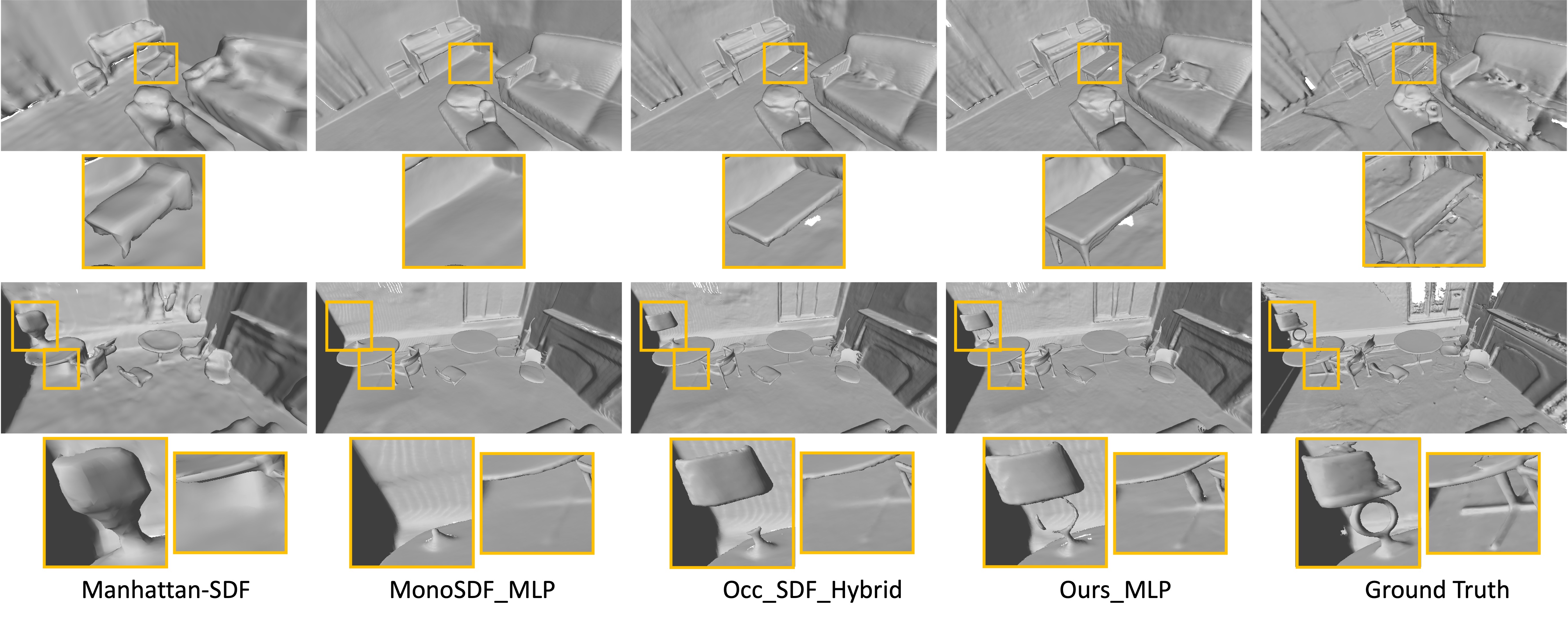}
  \caption{\textbf{Qualitative comparisons on ScanNet.} It can be seen that our method can reconstruct more surfaces, especially in thin regions.
  }
  \label{fig:sota_scannet}
\end{figure}

\begin{figure}[tb]
  \centering
  \includegraphics[width=0.93\linewidth]{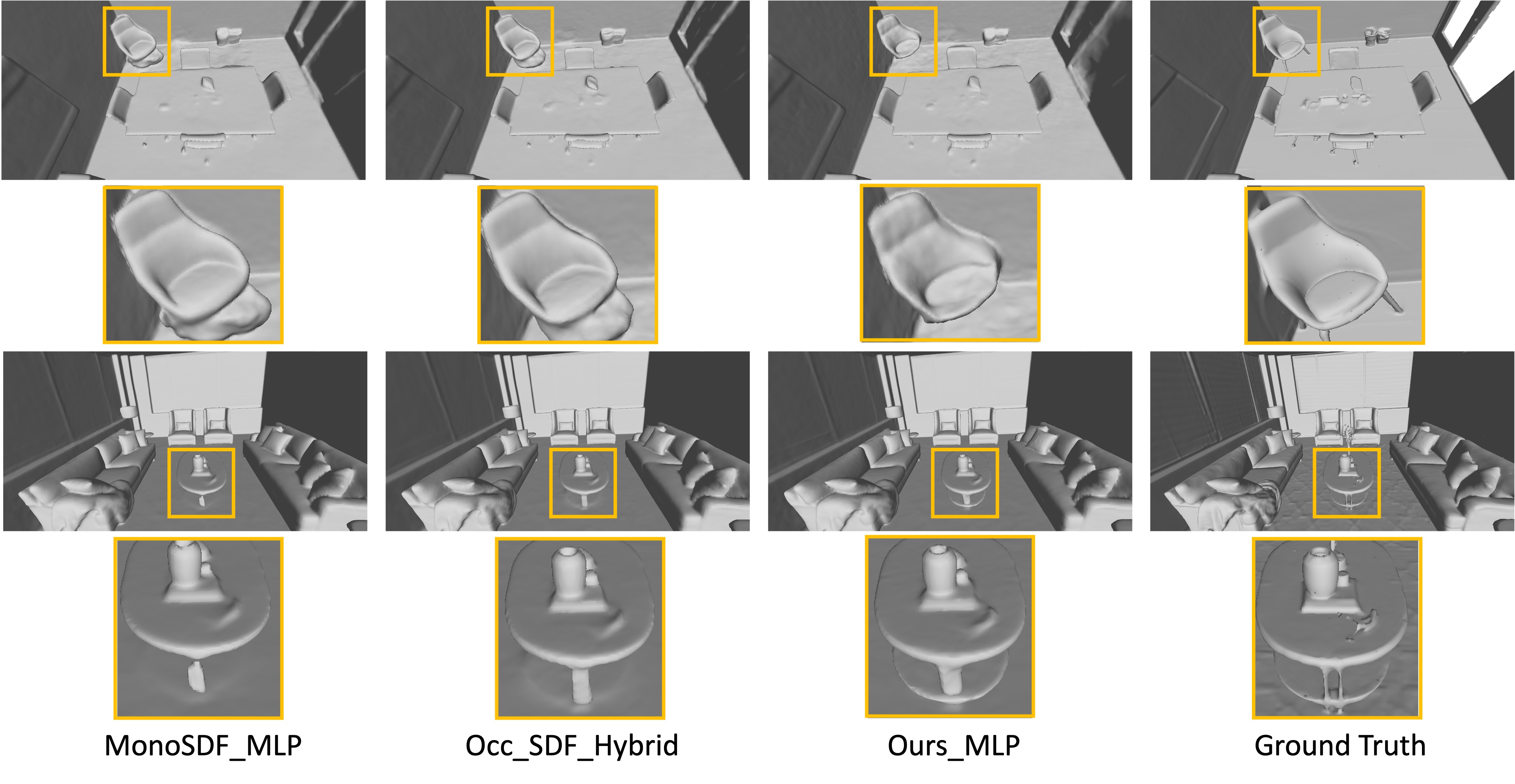}
  \caption{\textbf{Qualitative comparisons on Replica.} Compared to MonoSDF and Occ\_SDF\_Hybrid, our method can generate more accurate surfaces.
  }
  \label{fig:sota_replica}
\end{figure}

\begin{table}[tb]
\small
  \caption{Evaluation of 3D reconstruction meshes on Replica.
  }
  \label{tab:replica}
  \tabcolsep=5pt
  \centering

  \begin{tabular}{ccccc}
    \toprule
    &Method  & Normal C. $\uparrow$ & Chamfer-L1 $\downarrow$ &F-score $\uparrow$   \\\hline
  \multirow{2}{*}{Grid} &MonoSDF \cite{yu2022monosdf}  &  90.93 &3.23 &85.91 \\ 
  &Ours  &  \textbf{91.66} &\textbf{2.91}  &\textbf{88.79} \\ \hline
    \multirow{3}{*}{MLP} &MonoSDF \cite{yu2022monosdf}&  92.11 &2.94  &86.18 \\
  &Occ\_SDF\_Hybrid$^\dagger$  \cite{lyu2023learning}&  93.34 & \textbf{2.58} &{91.54} \\ 
    &Ours  &  \textbf{93.49} &{2.60}&\textbf{91.72} \\
    \bottomrule
  \end{tabular}
\end{table}

\begin{table}[tb]
\small
  \caption{Evaluation of 3D reconstruction meshes on the advanced set of Tanks and Temples. The F-score is reported in the table.
  }
  \label{tab:tnt}
  
  \tabcolsep=3pt
  \centering

  \begin{tabular}{ccccccc}
    \toprule
    &Method  & Auditorium & Ballroom & Courtroom & Museum &mean  \\\hline
    \multirow{3}{*}{Grid}&MonoSDF  \cite{yu2022monosdf}& 3.17   & 3.70 &13.75 &5.68 & 6.58 \\
    &Occ\_SDF\_Hybrid$^\dagger$ \cite{lyu2023learning} &3.76   &3.58  &14.04 &7.27 &7.16\\ 
    &Ours  &\textbf{4.84}   & \textbf{4.31} & \textbf{14.18} & \textbf{7.60} & \textbf{7.73}\\ \hline
   \multirow{3}{*}{MLP}& MonoSDF \cite{yu2022monosdf} & 3.09  & 2.47  & 10.00 & 5.10 & 5.17\\
    &Occ\_SDF\_Hybrid$^\dagger$  \cite{lyu2023learning}&3.21   &3.34  &10.29 &4.66 & 5.38\\ 
    &Ours  & \textbf{4.24} & \textbf{4.69} &\textbf{10.79}&\textbf{5.91} & \textbf{6.41}\\
    \bottomrule
  \end{tabular}
\end{table}

\begin{table}[tb]
\small
  \caption{Evaluation of the rendering quality. PSNR is given in this table.
  }
  \label{tab:psnr}
  
  \tabcolsep=7pt
  \centering

  \begin{tabular}{ccccc}
    \toprule
    & Method & {ScanNet}  & {Replica} & Tanks and Temples\\ \hline
    \multirow{2}{*}{Grid}&MonoSDF \cite{yu2022monosdf}  &  28.96 & 40.25  & 28.45  \\
    &Ours  &  \textbf{30.09}  & \textbf{42.04} & \textbf{29.52}  \\ \hline
    \multirow{3}{*}{MLP}&MonoSDF \cite{yu2022monosdf} &  26.40  & 34.45 & 24.13 \\
    &Occ\_SDF\_Hybrid$^\dagger$ \cite{lyu2023learning} &26.98    &35.50 & 24.72 \\
    &Ours  &  \textbf{27.77}  & \textbf{36.06} & \textbf{25.47} \\
    \bottomrule
  \end{tabular}
\end{table}

\noindent \textbf{Reconstruction on ScanNet \cite{dai2017scannet}.} \cref{tab:scannet} compares our method against other state-of-the-art (SOTA) methods on ScanNet. When contrasted with the baseline MonoSDF, our method consistently shows better performance across all metrics, regardless of whether utilizing MLP or grid representation. For example, employing grid representation, our method exhibits a 5.7\% higher F-score than MonoSDF. Using the MLP representation, our method outperforms MonoSDF by 1.0\% in recall, 9.8\% in precision, and 6.1\% in F-score. In addition, our method with the MLP representation surpasses other SOTA methods on almost all metrics, explaining the effectiveness of our method on reconstruction. Notably, our method achieves a 2.9\% higher F-score compared to Occ\_SDF\_Hybrid. 

\noindent \textbf{Reconstruction on Replica \cite{straub2019replica}.} The results on the synthetic dataset Replica are shown in \cref{tab:replica}. Using the grid representation, our method reports a notable improvement of 9.91\% in chamfer distance and 2.88\% in F-score. Similar increases are also illustrated in our method with the MLP representation across all metrics. Our approach also achieves on-par results with Occ\_SDF\_Hybrid. 

\noindent \textbf{Reconstruction on Tanks and Temples \cite{knapitsch2017tanks}.} \cref{tab:tnt} shows quantitative outcomes on the challenging large-scale dataset Tanks and Temples. The F-score for each scene - Auditorium, Ballroom, Courtroom, and Museum - along with their mean, is reported. Our method, with both the MLP and grid representations, outperforms Occ\_SDF\_Hybrid and MonoSDF. In particular, our F-score is 1.15\% higher (grid representation) and 1.24\% higher (MLP representation) than MonoSDF across all scenes.

\noindent \textbf{Visualization of Reconstruction Results.} \cref{fig:sota_scannet} and \cref{fig:sota_replica} report qualitative results on ScanNet and Replica respectively. In contrast to other methods, our approach can produce superior meshes and recall more regions, particularly for thin regions/small objects, such as chair legs and table legs. 
Although our method relies on the SRDF for improved density in volume rendering while the surface geometry is extracted by the SDF, both the proposed SRDF-SDF consistency loss and self-supervised visibility task guide the optimization of the SDF to be more consistent with the SRDF, resulting in more accurate surface boundary. Additional visualization is provided in the supplementary material.

\noindent \textbf{View Synthesis.} 
\cref{tab:psnr} also provides a comparison of view synthesis performance on training views. Notably, 2D views are rendered with density derived from SRDF. On the three datasets, our method outperforms the baseline and Occ\_SDF\_Hybrid in rendering training views. Overall, our method not only enhances surface reconstruction but also generates more accurate views. The visualization can be found in the supplementary material.

\subsection{Ablation Study}
\label{sec:exp ab}

\begin{table}[tb]
\small
  \caption{Ablation study of our method on ScanNet.
  }
  \label{tab:ablation}
  \tabcolsep=2.5pt
  \centering

  \begin{tabular}{ccccccc}
    \toprule
    & Method  & Acc $\downarrow$ & Comp $\downarrow$ &Prec $\uparrow$ &Recall $\uparrow$ &F-score $\uparrow$   \\
    \hline
    a& Baseline & \textbf{0.035} &0.048 &0.799 &0.681 &0.733 \\
    b& RS. Den. &  {0.040} &0.044 &0.772 &0.720 &0.745 \\
    c&RS. Den. + Con. L. &  {0.039} &{0.041} &{0.794} &{0.760} &{0.776} \\ \hline 
    d&RS. Den. + Con. L. + Vis.(SDF) &  {0.039} &\textbf{0.040}  &\textbf{0.809} &{0.772} &{0.789} \\
    e&RS. Den. + Con. L. + Vis.(SRDF) &  {0.042} &\textbf{0.040} &{0.804} &{0.774} &{0.788} \\
    f&RS. Den. + Con.L. + Vis.   &  {0.040} &\textbf{0.040} &\textbf{0.809} &\textbf{0.779} &\textbf{0.794} \\
    \bottomrule
  \end{tabular}
\end{table}

\begin{figure}[tb]
  \centering
  \includegraphics[width=0.94\linewidth]{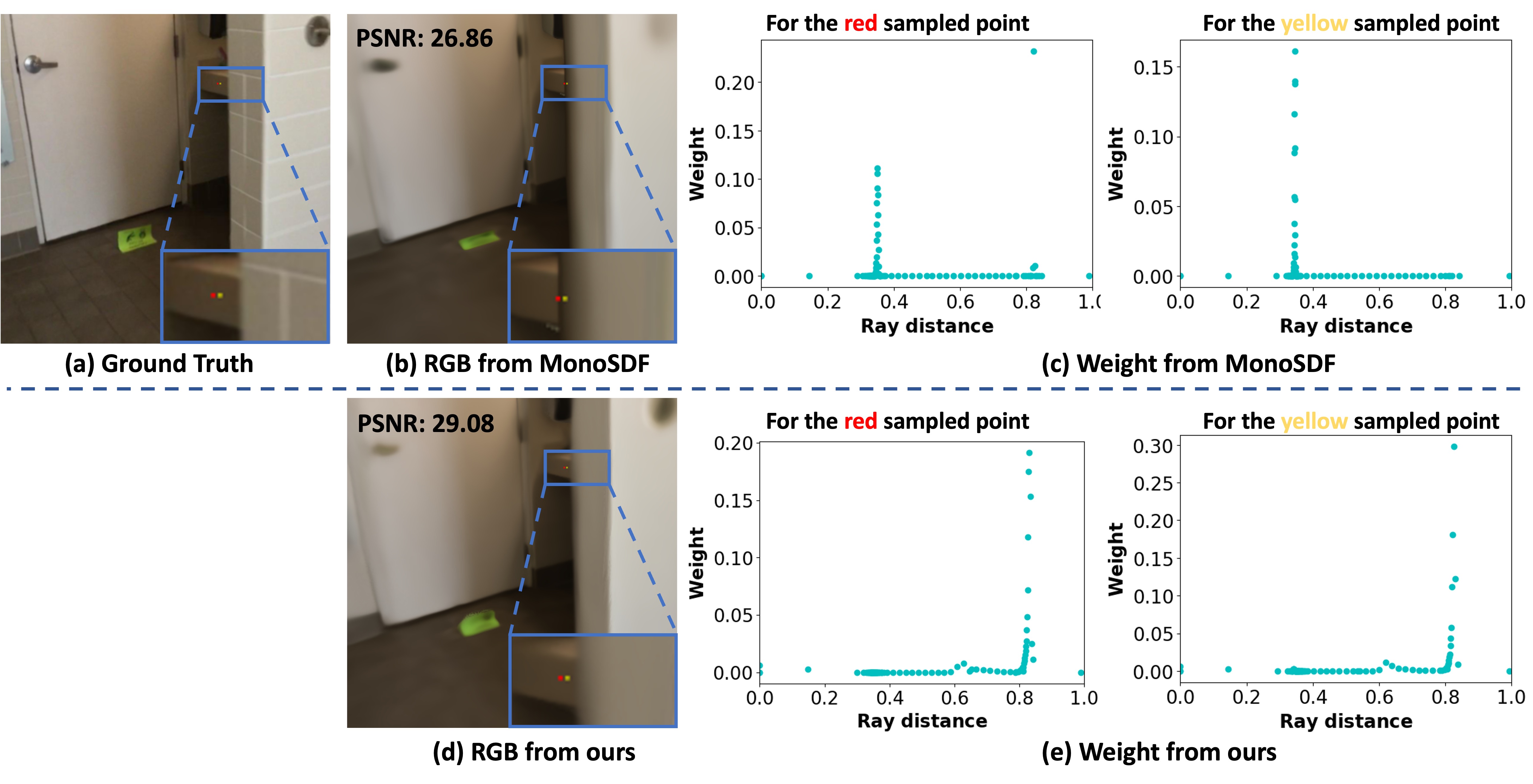}
  \caption{\textbf{Comprison of the yielded weight and image.} The weight and image generated by our approach align more closely with the actual observations.
  }
  \label{fig:weight}
\end{figure}

\cref{tab:ablation} presents an ablation study to evaluate the effectiveness of our design on ScanNet with MLP representation.
The ablation experiments consist of: (a) \emph{Baseline}: MonoSDF. (b) \emph{RS. Den.}: this structure models the volume density using the ray-specific SRDF, without the SRDF-SDF consistency loss and self-supervised visibility task. (c) \emph{RS. Den. + Con. L.}: Building upon the structure in (b), this setup adopts the proposed SRDF-SDF consistency loss. (d) \emph{RS. Den. + Con. L. + Vis.(SDF)}: based on (c), this configuration adds the self-supervised visibility task, in which only SDF is used to compute the visibility label. (e) \emph{RS. Den. + Con. L. + Vis.(SRDF)}: Different from structure (d), this method relies solely on SRDF to compute the visibility label. (f) \emph{RS. Den. + Con. L. + Vis.}: this structure represents the comprehensive model of our method, wherein the visibility labels integrate priors from both SRDF and SDF predictions.

The comparison between configurations (a) and (b) highlights that our design using SRDF to model the density, without any additional constraints, results in a better performance compared to MonoSDF. The SRDF-SDF consistency loss in (c) surpasses the structure in (b) by 3.1\% in F-score. This shows the significance of sign consistency. Compared to (c), the self-supervised visibility task in (d)-(f) enhances the reconstruction performance. In particular, structure (f) obtains a better F-score than structures (d) and (e). These explain the effectiveness of our self-supervised visibility task and the computation of visibility labels integrating both SRDF and SDF priors. The visualization of the ablation study is given in the supplementary material.

Furthermore, \cref{fig:weight} compares the weight distribution for the red and yellow sampled points generated by MonoSDF and our method. (1) The analysis of the yellow sampled point demonstrates the benefits of our method for reconstruction: MonoSDF generates high weights and negative SDF values when sampled rays are near the white wall, resulting in false surfaces that significantly deviate from the ground truth. In contrast, our method yields more accurate surfaces and thus enhances both completeness and recall. (2) The analysis of the red sampled point explains the reason why our method can generate accurate views: MonoSDF produces weights with two local maxima, causing an inaccurate RGB value. This occurs because the SDF along the camera ray fluctuates, resulting in varying density and weight. In comparison, our method produces uni-modal weights that peak only around the surface, resulting in more accurate views. The quantitative PSNR in \cref{fig:weight}{b} and \cref{fig:weight}{d} confirms that our rendered image is more accurate compared to MonoSDF. These findings align with our motivation to achieve a more consistent and accurate weight distribution, thereby reconstructing better surfaces and generating high-quality views. This underscores the effectiveness of using ray-specific SRDF to model volume density in our approach.

\section{Conclusion}
\label{sec:conclusion}
This work proposes a novel method for neural indoor scene reconstruction, modeling density function with ray-specific SRDF. Firstly, this study analyzed that using the SDF to parameterize the volume density may introduce noises in multi-object indoor scenes, and the SRDF exhibits stronger relationships with the actual 2D observations. Motivated by this, this work employed SRDF to model the volume density, while SDF mainly focuses on the representation of the 3D surface. Secondly, a SRDF-SDF consistency loss has been introduced to align the sign between SRDF and SDF. Thirdly, a self-supervised visibility task has been designed to distinguish whether the sampled 3D point is physically visible or occluded. It combined the prior from SDF and SRDF as visibility labels to improve geometry reconstruction without any other annotations. Experimental results on real-world and synthetic datasets showed that our method excels not only in reconstructing better surfaces but also in generating more accurate views.

%
%
\bibliographystyle{splncs04}
\bibliography{main}

\clearpage
\vspace{5mm}
{\Large\centering \textbf{Supplementary Material}} 
\vspace{5mm}
\appendix

\renewcommand*{\thefigure}{S\arabic{figure}}
\renewcommand*{\thetable}{S\arabic{table}}
\renewcommand*{\theequation}{S-\arabic{equation}}
\setcounter{table}{0}
\setcounter{figure}{0}
\setcounter{equation}{0}

This supplementary document provides additional details and experimental results of our approach. The definitions of evaluation metrics are given in Sec. \ref{sec:metrics}.  Sec. \ref{sec:loss} details loss functions employed in our method. The implementation details are discussed in Sec. \ref{sec:network}. Sec. \ref{sec:extra_qualitative} shows additional qualitative and quantitative results. Sec. \ref{sec:analysis} offers a complementary analysis of the efficacy of our approach.

\section{Evaluation Metrics}
\label{sec:metrics}
The definitions of metrics are given in \cref{tab:metrics}, including accuracy (acc), completeness (comp), precision (prec), recall, F-score, normal consistency (normal c.), chamfer L1 distance (Chamfer-L1), and peak signal-to-noise ratio (PSNR). Lower values denote superior performance for accuracy, completeness, and chamfer L1 distance, whereas higher values are preferable for precision, recall, F-score, normal consistency, and PSNR.
\begin{table}[!h]
  \caption{Definitions of evaluation metrics. $\textbf{p}$ is the sampled point cloud from the predicted mesh, while $\bar{\textbf{p}}$ is from the ground truth. $\textbf{n}_{\textbf{p}}$ and $\textbf{n}_{\bar{\textbf{p}}}$ represent normal vectors computed from the prediction and ground truth respectively. $\textbf{C}_{pred}^n$ / $\textbf{C}_{gt}^n$ is the \emph{n}-th predicted/ground truth image. $N_I$ is the number of images for evaluation.
  }
  \label{tab:metrics}
  \tabcolsep=10pt
  \centering
  \renewcommand{\arraystretch}{1.1}
  \begin{tabular}{lc}
    \toprule
    Metric & Definition \\ \hline
  Acc&    $\text{mean}_{\textbf{p} \in {{P}}}(\text{min}_{\bar{\textbf{p}} \in {\bar{P}}}||\textbf{p}-\bar{\textbf{p}}||_1)$ \\ 
  Comp&   $\text{mean}_{\bar{\textbf{p}} \in  {\bar{P}}}(\text{min}_{\textbf{p} \in {{P}}}||\textbf{p}-\bar{\textbf{p}}||_1)$    \\ 
  Chamfer-L1 & $\frac{ \text{Acc} + \text{Comp}}{2}$ \\
  Prec& $\text{mean}_{\textbf{p} \in {{P}}}(\text{min}_{\bar{\textbf{p}} \in  \bar{P}}||\textbf{p}-\bar{\textbf{p}}||_1<.05)$  \\ 
  Recall& $\text{mean}_{\bar{\textbf{p}} \in \bar{P}}(\text{min}_{\textbf{p} \in {{P}}}||\textbf{p}-\bar{\textbf{p}}||_1<.05)$ \\
  F-score & $\frac{2\times \text{Prec} \times \text{Recall}}{\text{Prec} + \text{Recall}}$ \\
   Normal-Acc & $\operatorname{mean}_{\textbf{p} \in P}\left(\textbf{n}_{\textbf{p}}^T \textbf{n}_{\bar{\textbf{p}}}\right) \text { s.t. } \bar{\textbf{p}}=\underset{\bar{\textbf{p}} \in \bar{P}}{\operatorname{argmin}}\left\|\textbf{p}-\bar{\textbf{p}}\right\|_1$ \\
   Normal-Comp & $\operatorname{mean}_{\bar{\textbf{p}} \in \bar{P}}\left(\textbf{n}_{\textbf{p}}^T \textbf{n}_{\bar{\textbf{p}}}\right) \text { s.t. } \textbf{p}=\underset{\textbf{p} \in P}{\operatorname{argmin}}\left\|\textbf{p}-\bar{\textbf{p}}\right\|_1$ \\
  Normal Consistency & $\frac{\text{Normal-Acc} + \text{Normal-Comp}}{2}$\\
  PSNR & $\text{mean}_{n \in [1, N_{I}]}(20 \cdot \text{log}_{10}(1/\sqrt{\text{MSE}(\textbf{C}_{pred}^n, \textbf{C}_{gt}^n)})$) \\
    \bottomrule
  \end{tabular}
\end{table}

\section{Details of Loss Function}
\label{sec:loss}
This section provides the computation for the RGB loss $\mathcal{L}_{c}$, the depth loss $\mathcal{L}_d$, the normal loss $\mathcal{L}_n$, the smooth loss $\mathcal{L}_s$, and the Eikonal loss $\mathcal{L}_e$.

To enhance the prediction of geometry and appearance, the RGB loss comprises the color loss with density derived from SRDF, denoted as $\mathcal{L}_{c}^{\text{SRDF}}$, and another color loss with density transformed from SDF, denoted as $\mathcal{L}_{c}^{\text{SDF}}$. They are calculated as follows:
\begin{equation}
\begin{aligned}
\mathcal{L}_{c} &= \mathcal{L}_{c}^{\text{SRDF}} + \mathcal{L}_{c}^{\text{SDF}} \\
\mathcal{L}_{c}^{\text{SRDF}} &= \frac{1}{N_s}\sum_{\rvr \in \mathcal{S}^2} {\parallel \textbf{C}^{\text{SRDF}} - \bar{\textbf{C}}\parallel}_1  \\
\mathcal{L}_{c}^{\text{SDF}} &= \frac{1}{N_s}\sum_{\rvr \in \mathcal{S}^2} {\parallel \textbf{C}^{\text{SDF}} - \bar{\textbf{C}}\parallel}_1 \\
\end{aligned}
\end{equation}
where $\mathcal{S}^2$ represents the set of rays in a minibatch. $\bar{\textbf{C}}$ denotes the ground truth of color. $\textbf{C}^{\text{SRDF}}$ is the rendered color with density derived from SRDF as detailed in Eq. ({5}), while $\textbf{C}^{\text{SDF}}$ represents the color rendered with the density that is derived from SDF according to Eq. ({3}). $N_s$ refers to the number of points sampled per minibatch. 

The depth and normal losses, serving to constrain the 3D geometry, are defined as follows:

\begin{equation}
\begin{aligned}
{D} &= \sum_{i=1}^N T_i\alpha_i {z}_i,  {\textbf{N}}=\sum_{i=1}^N T_i\alpha_i \mathbf{n}_i \\
\mathcal{L}_d &= \frac{1}{N_s}\sum_{\rvr \in \mathcal{S}^2} {\parallel(w{{D}}+q) -\bar{D}\parallel}_2 \\
\mathcal{L}_n &= \frac{1}{N_s}\sum_{\rvr \in \mathcal{S}^2} ({\parallel {\textbf{N}} - \bar{\textbf{N}}\parallel}_1 + {\parallel 1 - {\textbf{N}}^T\bar{\textbf{N}}\parallel}_1) \\
\end{aligned}
\end{equation}
where $\bar{D}$ and $\bar{\textbf{N}}$ correspond to the ground truth of depth and normal, predicted by the pre-trained Omnidata model \cite{eftekhar2021omnidata}. ${{D}}$ and ${\textbf{N}}$ denote rendered depth and normal, respectively. $T_i$ and $\alpha_i$ are computed by Eq. ({1}) with density derived from SRDF. To address potential depth ambiguities, a scale parameter $w$ and a shift parameter $q$ are utilized to ensure predicted depth ${D}$ matches the ground truth $\bar{D}$, as utilized by MonoSDF.

The smooth loss, defined in \cref{eq: smooth loss}, is computed by considering the gradients of the SDF between adjacent points. This encourages a smoother reconstructed surface.
\begin{equation}
\begin{aligned}
\mathcal{L}_s = \frac{1}{N_{sm}}\sum_{\textbf{p} \in \mathcal{X}} {\parallel \nabla{d_\Omega(\textbf{p})} - \nabla{d_\Omega(\textbf{p} + \varepsilon)}\parallel}_2 \\
\end{aligned}
\label{eq: smooth loss}
\end{equation}
where $\varepsilon$ is a small perturbation applied to the point. $\mathcal{X}$ is the set of sampled points. $d_\Omega(\textbf{p})$ refers to the SDF output produced by the geometry MLP. Additionally, $N_{sm}$ indicates the number of points for computing the smooth loss. 

The Eikonal loss \cite{gropp2020implicit} serves as a regularization term for SDF prediction, detailed in \cref{eq: eikonal loss}.
\begin{equation}
\begin{aligned}
\mathcal{L}_e = \frac{1}{N_{ei}}\sum_{\textbf{p} \in \mathcal{X}} ({\parallel \nabla{d_\Omega(\textbf{p})}\parallel}_2 - 1)^2 \\
\end{aligned}
\label{eq: eikonal loss}
\end{equation}
where $N_{ei}$ refers to the number of sampled points for the Eikonal loss.

\section{Implementation Details}
\label{sec:network}

The MLP representation employs a geometry MLP comprising 8 layers with a hidden dimension of 256, while the grid representation utilizes two layers. Both the color MLP and SRDF MLP consist of two layers each, with an intermediate output comprising 256 channels. Softplus is used as the activation layer in the geometry MLP and SRDF MLP, while ReLU is applied in the color MLP. The network utilizes geometric initialization \cite{atzmon2020sal} and is optimized over 200,000 iterations with Adam optimizer \cite{kingma2014adam}. The resolution of 2D output is set to $384 \times 384$. Each iteration includes the sampling of 1024 rays. The initial learning rate is configured as 5e-4 for geometry MLP and color MLP, 1e-5 for the SRDF branch, and 1e-2 for feature grids. The hyperparameter $k$ in the SRDF-SDF consistency loss is set to 12. Loss weights $\{\lambda_d, \lambda_n, \lambda_{e}, \lambda_s, \lambda_{con}, \lambda_{vis}\}$  are \{0.1, 0.05, 0.05, 0.005, 1.0, 0.001\}. Following MonoSDF, the ground truth for monocular geometric cues is predicted by the pre-trained Omnidata model \cite{eftekhar2021omnidata}. The optimization is performed on one NVIDIA A100 GPU.

\section{Additional Results}
\label{sec:extra_qualitative}

\subsection{Additional Ablation Study}
\noindent \textbf{Impact of SRDF Branch:} Additional ablation study, shown in \cref{tab:srdf}, evaluates the importance of SRDF branch. The experiments include: (a) \emph{Baseline}: MonoSDF with MLP representation. (b) \emph{SDF to SRDF}: this setup generates SRDF from the predicted SDF, eliminating the need for the SRDF branch. The SRDF is then used to calculate volume density. (c) \emph{SRDF branch:} SRDF is predicted by the SRDF branch. In \cref{tab:srdf}, Row {(b)} performs much worse than the baseline. The reason is that the SDF prediction for a single point is affected by all nearby objects, which can result in inaccuracies, generating false surfaces and imprecise SRDF. Thus, without the SRDF branch, SRDF-based volume rendering may yield inaccurate weights, encountering the same issues as SDF-based volume rendering. Moreover, since the surface is interpolated by the SDF of near-surface points, SRDF derived from SDF lacks adequate supervision for SDF during training, which can result in poorer geometry. Hence, generating SRDF from the SDF branch is not optimal. Instead, Row {(c)} shows that the structure with an SRDF branch can avoid the influence of inaccurate SDF and achieve better results. Additionally, the joint optimization of SRDF and SDF branches can integrate SRDF information into the SDF branch, thereby improving SDF prediction. In conclusion, the SRDF branch is crucial for SRDF-based volume rendering.

\begin{figure}[tb]
  \centering
  \includegraphics[width=0.8\linewidth]{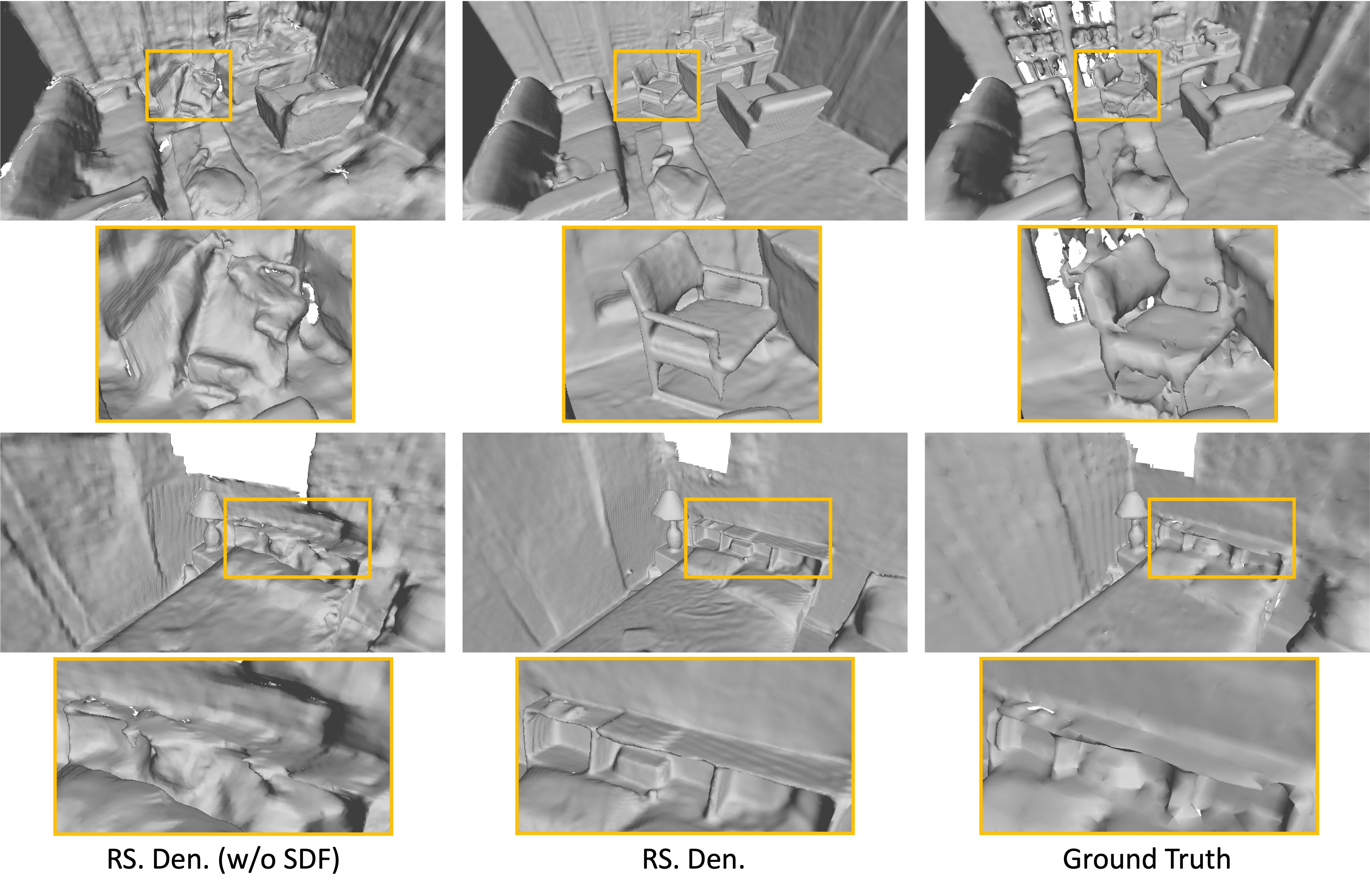}
  \caption{\textbf{Visualization of structures with/without SDF on ScanNet.} With SDF representing the surface geometry, our method (\emph{RS. Den.}) generates smoother and more accurate surfaces.
  }
  \label{fig:ab_onlysrdf}
\end{figure}

\begin{table}[tb]
  \caption{Ablation study for the impact of SDF in our method on ScanNet.
  }
  \label{tab:onlysrdf}
  \tabcolsep=5pt
  \centering
  
  \begin{tabular}{cccccc}
    \toprule
    Method  & Acc $\downarrow$ & Comp $\downarrow$ &Prec $\uparrow$ &Recall $\uparrow$ &F-score $\uparrow$   \\
    \hline
    RS. Den. (w/o SDF) &  {0.061} &{0.079} &{0.600} &{0.540} &{0.568} \\
    RS. Den. &  {0.040} &0.044 &0.772 &0.720 &0.745 \\
    \bottomrule
  \end{tabular}
\end{table}

\begin{table}[tb]
  \caption{Ablation study for the impact of SRDF branch in our method on ScanNet.
  }
  \label{tab:srdf}
  \tabcolsep=6pt
  \centering

  \begin{tabular}{ccccccc}
    \toprule
    &Method  & Acc $\downarrow$ & Comp $\downarrow$ &Prec $\uparrow$ &Recall $\uparrow$ &F-score $\uparrow$   \\
    \hline
   a &Baseline &  {0.035} &0.048 &0.799 &0.681 &0.733 \\ 
   b &  SDF to SRDF &  {0.049} &0.068 &0.673 &0.591 &0.628 \\
   c & SRDF branch &  {0.040} & {0.044} & {0.772} & {0.720} & {0.745} \\
    \bottomrule
  \end{tabular}
\end{table}

\noindent \textbf{Impact of SDF Prediction:} \cref{tab:onlysrdf} measures the importance of using SDF for 3D surface localization. Two configurations are compared: (1) \emph{RS. Den. (w/o SDF):} This structure only predicts ray-specific SRDF, without using SDF for 3D geometry representation. During training, color loss and depth loss are used to optimize the network. Since SRDF relies on viewing direction, directly extracting the surface from SRDF is not feasible. Therefore, during inference, to generate 3D meshes, the network first renders 2D depth with volume density derived from SRDF, after which TSDF Fusion \cite{curless1996volumetric} is adopted to fuse the multi-view depth maps. (2) \emph{RS. Den.:} This structure predicts both SRDF and SDF, where SRDF models the volume density and SDF locates 3D surfaces. Notably, the SRDF-SDF consistency loss and self-supervised visibility task are not employed in this ablation study. In \cref{tab:onlysrdf}, The structure \emph{RS. Den. (w/o SDF)} achieves much worse results than \emph{RS. Den.}. This occurs because the depth label in the depth loss is generated by a pre-trained network, which may yield inaccurate absolute depth and introduce noises. Besides, the scale inconsistency between overlapped regions in the predicted depth also results in coarse surfaces, \eg walls and floors, as illustrated in \cref{fig:ab_onlysrdf}. In contrast, our structure with SDF yields superior reconstruction meshes.

\begin{figure}[tb]
  \centering
  \includegraphics[width=0.9\linewidth]{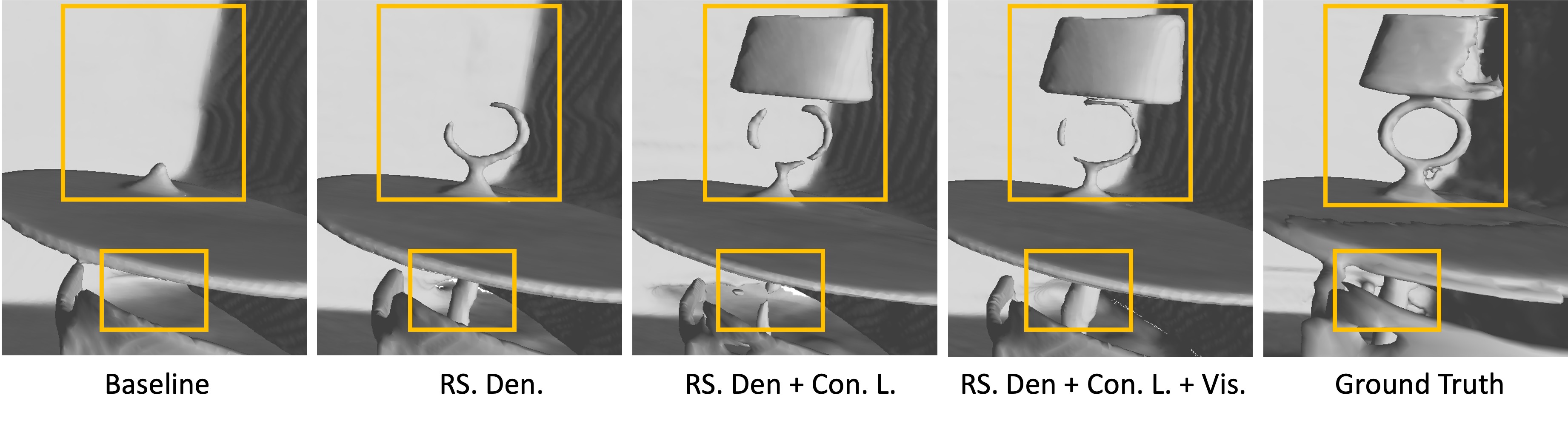}
  \caption{\textbf{Visualization for ablation study.} Our ray-specific volume density (\emph{RS. Den.}), SRDF-SDF consistency loss (\emph{Con. L.}), and self-supervised visibility task (\emph{Vis.}) all lead to better performance.
  }
  \label{fig:ab}
\end{figure}

\noindent \textbf{Visualization for Ablation Study:} To further comprehend the effectiveness of our proposed method, \cref{fig:ab} shows the visualization for ablation study. It can be seen that our ray-specific volume density, SRDF-SDF consistency loss, and self-supervised visibility task all contribute to high-quality reconstruction.

\subsection{Applying Our SRDF-based Solution to NeuRIS~\cite{wang2022neuris}}

\begin{table}[tb]
  \caption{Evaluation of 3D reconstruction meshes on ScanNet. 
  }
  \label{tab:neuris}
  \tabcolsep=5pt
  \centering

  \begin{tabular}{cccccc}
    \toprule
    Method  & Acc $\downarrow$ & Comp $\downarrow$ &Prec $\uparrow$ &Recall $\uparrow$ &F-score $\uparrow$   \\
    \hline
    NeuRIS \cite{wang2022neuris} &  {0.050} &{0.049} &{0.717} &{0.669} &{0.692} \\
    NeuRIS + ours &  {0.044} &0.046 &0.768 &0.717 &0.742 \\
    \bottomrule
  \end{tabular}
\end{table}

\begin{figure}[tb]
  \centering
  \includegraphics[width=0.9\linewidth]{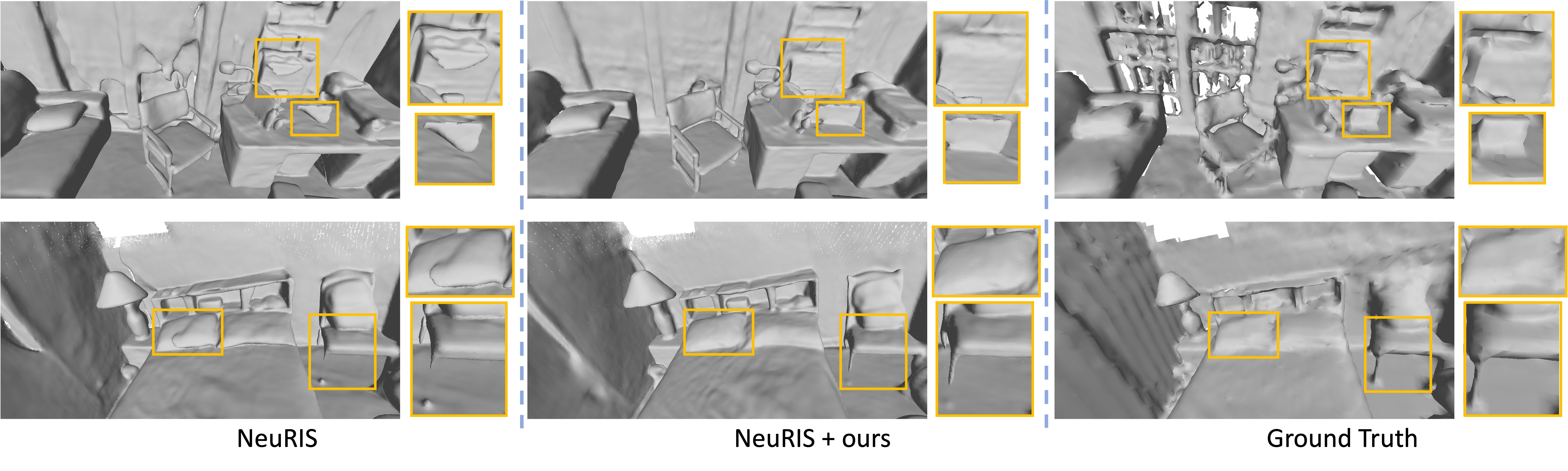}
  \caption{\textbf{Visualization of NeuRIS and our method on ScanNet.} Our method improves the reconstruction performance.
  }
  \label{fig:neuris}
\end{figure}

In 3D surface reconstruction, VolSDF \cite{yariv2021volume} and NeuS \cite{wang2021neus} are two mainstream baselines that apply volume rendering based on SDF. Our baseline MonoSDF, discussed in the main text, is an extension of VolSDF for indoor scene reconstruction. To assess the generalization of our method, we also apply our method to the NeuS-based method, namely NeuRIS \cite{wang2022neuris}, which also operates on indoor scenes. \cref{tab:neuris} compares our method based on NeuRIS and the baseline NeuRIS. Our approach outperforms NeuRIS by a significant margin, \eg achieving a 4.8\% increase in recall and a 5.0\% increase in F-score. Qualitative comparisons are given in \cref{fig:neuris}. Our approach demonstrates the ability to reconstruct more accurate surfaces compared to NeuRIS. Overall, these results highlight the efficacy of our method across both VolSDF-based and NeuS-based methods.

\subsection{Additional Reconstruction Results}

\begin{table}[tb]

\caption{3D reconstruction metrics for objects on ScanNet.}
\label{tab: object}
\setlength\tabcolsep{10pt}

\centering
\begin{tabular}{c|cc|cc}
\hline
& \multicolumn{2}{c}{3-class objects} & \multicolumn{2}{c}{All objects} \\ \hline
 & Comp$\downarrow$ & Recall$\uparrow$  &  Comp$\downarrow$ & Recall$\uparrow$\\ \hline

{MonoSDF\_Grid}& 0.041&  0.754 & 0.047   &  0.662      \\ 
{Ours\_Grid} & {0.035} & {0.801} & {0.036} & {0.779}        \\ \hline
{MonoSDF\_MLP}  & 0.034 & 0.849   &    0.038 & 0.772   \\ 
{Ours\_MLP}  &\textbf{0.026}  & \textbf{0.886}  & \textbf{0.031} &\textbf{0.827}   \\ \hline

\end{tabular}

\end{table}

To further verify that our method is beneficial for small/thin objects, we also evaluate the reconstruction performance for objects. Using the semantic segmentation mask as prior, we adopt two evaluation approaches: (1) Assessing `\emph{3-class objects}' including chairs, tables, and lamps, which often contain thin/small regions. (2) Evaluating `\emph{all objects}' excluding walls, ceilings, floors, windows, and doors, as they may not include thin/small regions.  In \cref{tab: object}, compared to MonoSDF, our method demonstrates significant improvements in both completeness and recall, confirming its effectiveness in object reconstruction. 

\begin{figure}[tb]
  \centering
  \includegraphics[width=0.9\linewidth]{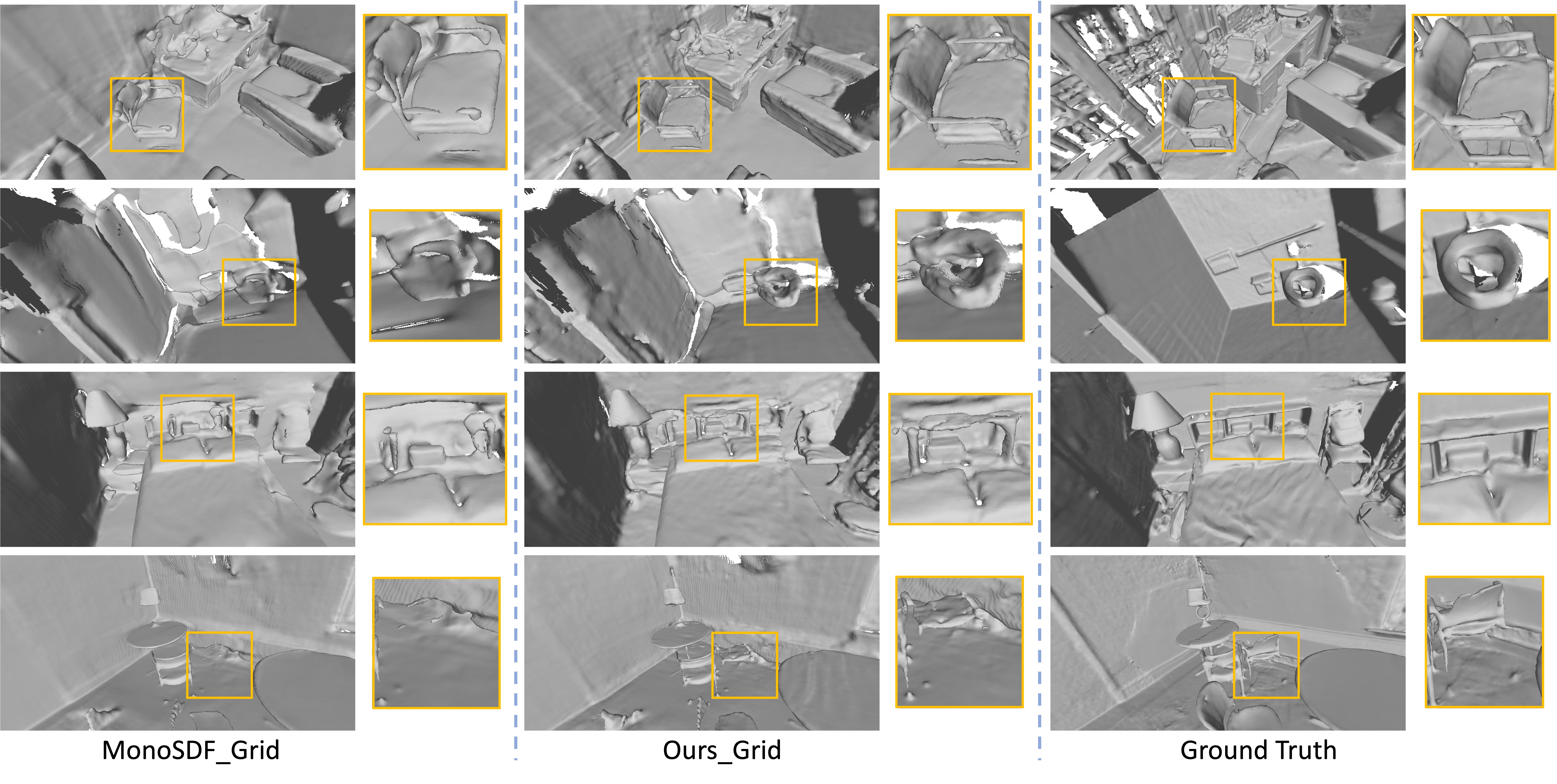}
  \caption{\textbf{Additional results on ScanNet.} Our method can recall more regions.
  }
  \label{fig:scannet_mesh}
\end{figure}

\begin{figure}[tb]
  \centering
  \includegraphics[width=0.9\linewidth]{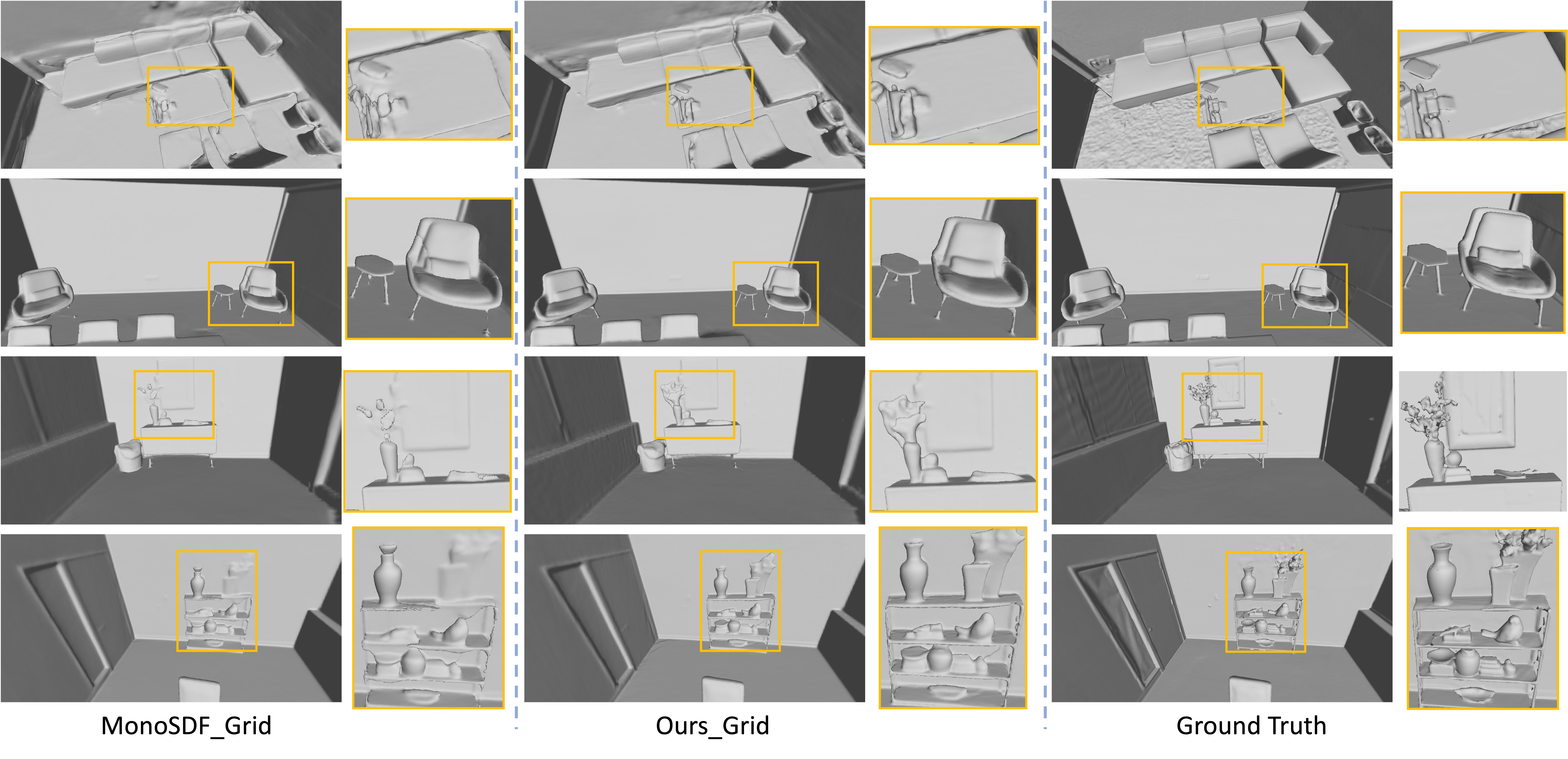}
  \caption{\textbf{Additional results on Replica.} Our method can generate more details.
  }
  \label{fig:replica_mesh}
\end{figure}

\cref{fig:scannet_mesh} and \cref{fig:replica_mesh} give additional visualizations on ScanNet and Replica, respectively, with the grid representation. In comparison to MonoSDF, our method demonstrates superior performance in capturing more surfaces and details, \eg chairs, toilets, and shelves.

\begin{figure}[tb]
  \centering
  \includegraphics[width=0.89\linewidth]{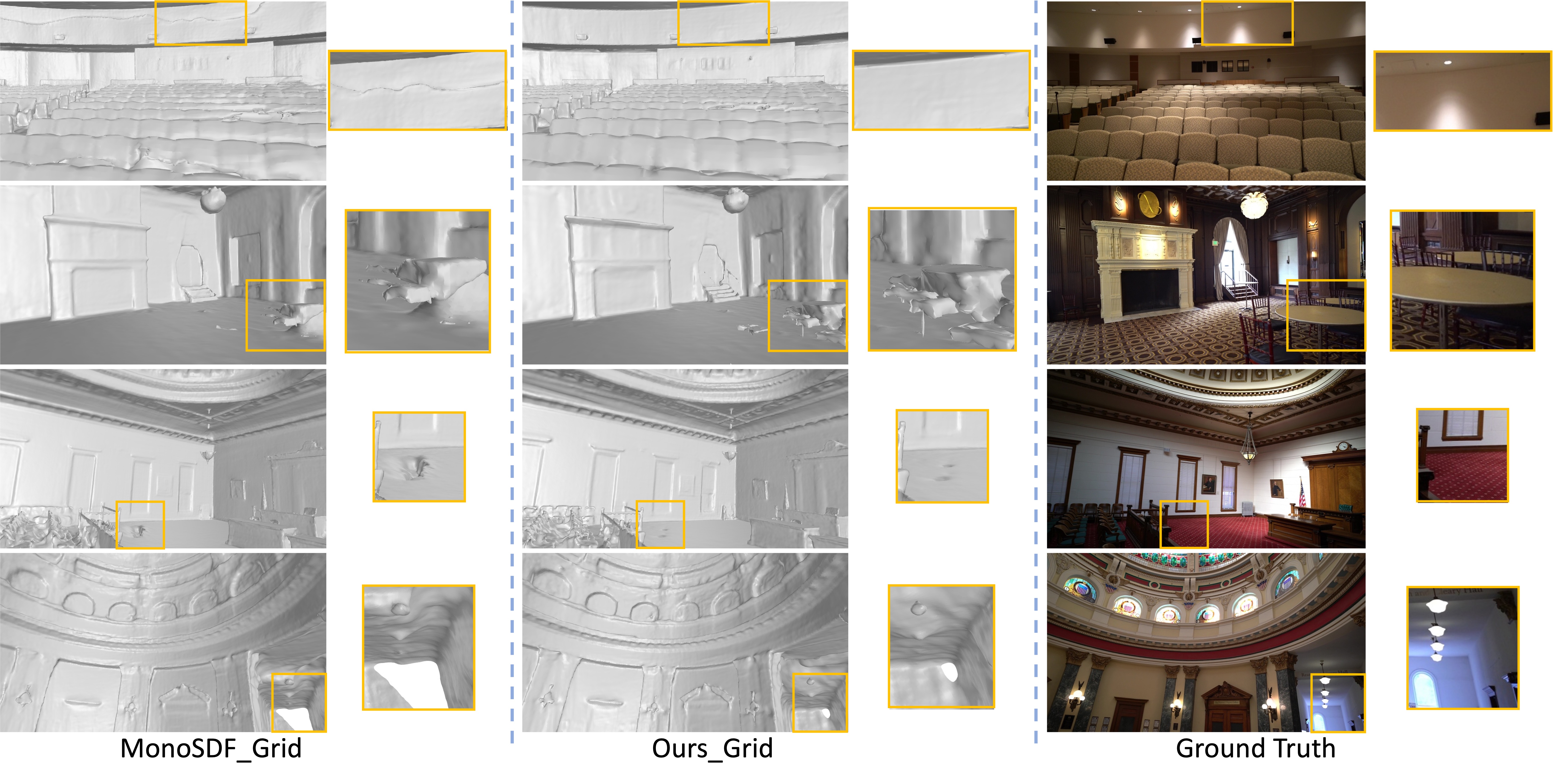}
  \caption{\textbf{Additional results on Tanks and Temples.} Our method enhances the quality of reconstructed meshes.
  }
  \label{fig:tnt_mesh}
\end{figure}

\begin{figure}[!h]
  \centering
  \includegraphics[width=1.0\linewidth]{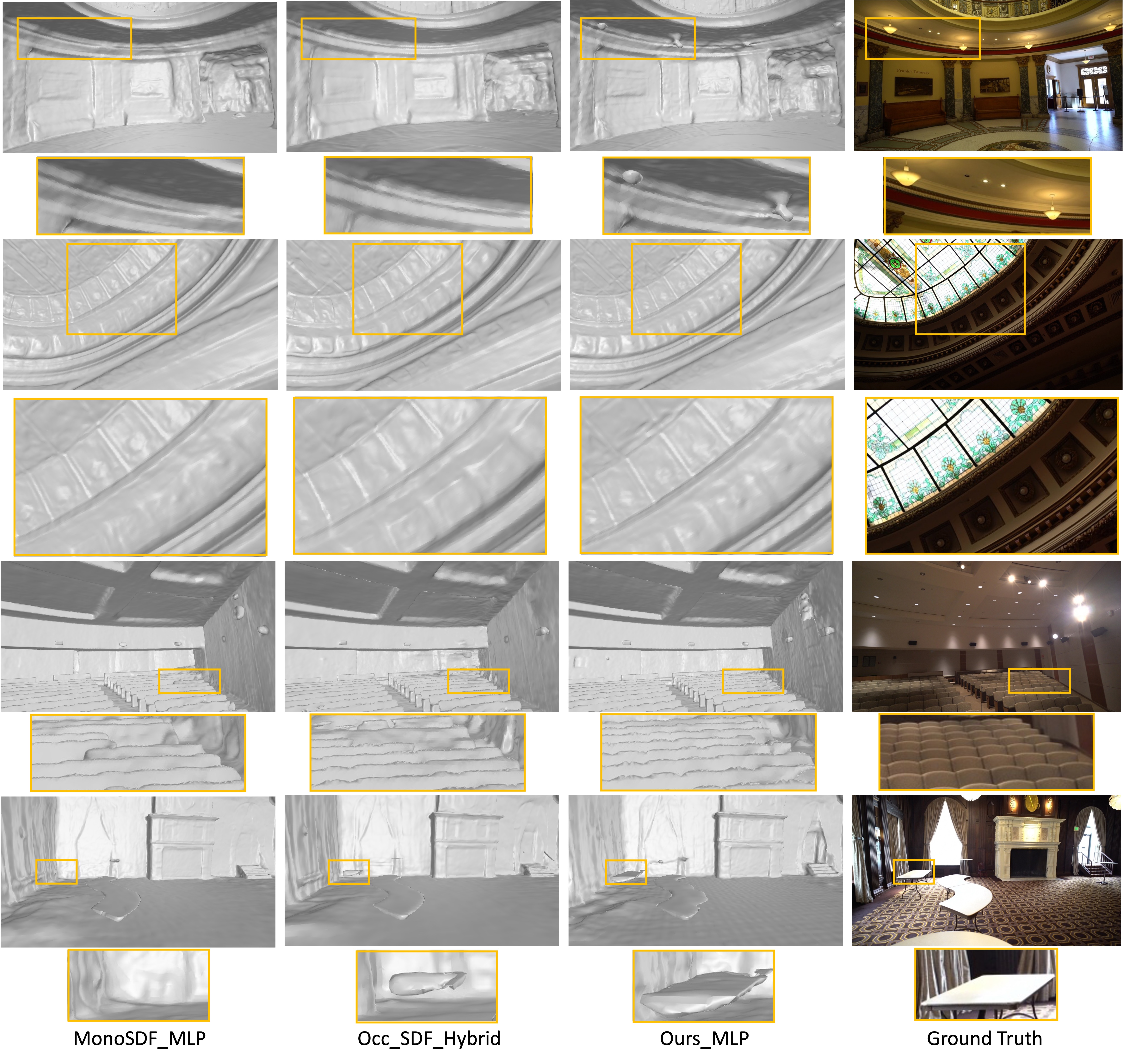}
  \caption{\textbf{Qualitative comparisons on Tanks and Temples.} Compared to other methods, our method can reconstruct more surfaces and recover details.
  }
  \label{fig:sota_tnt}
\end{figure}

\cref{fig:tnt_mesh} and \cref{fig:sota_tnt} present the reconstructed surfaces using grid and MLP representations on Tanks and Temples. It can be seen that our approach yields much better surfaces compared to MonoSDF and Occ\_SDF\_Hybrid.

\subsection{Visualization for View Synthesis}

\begin{figure}[!h]
  \centering
  \includegraphics[width=1.0\linewidth]{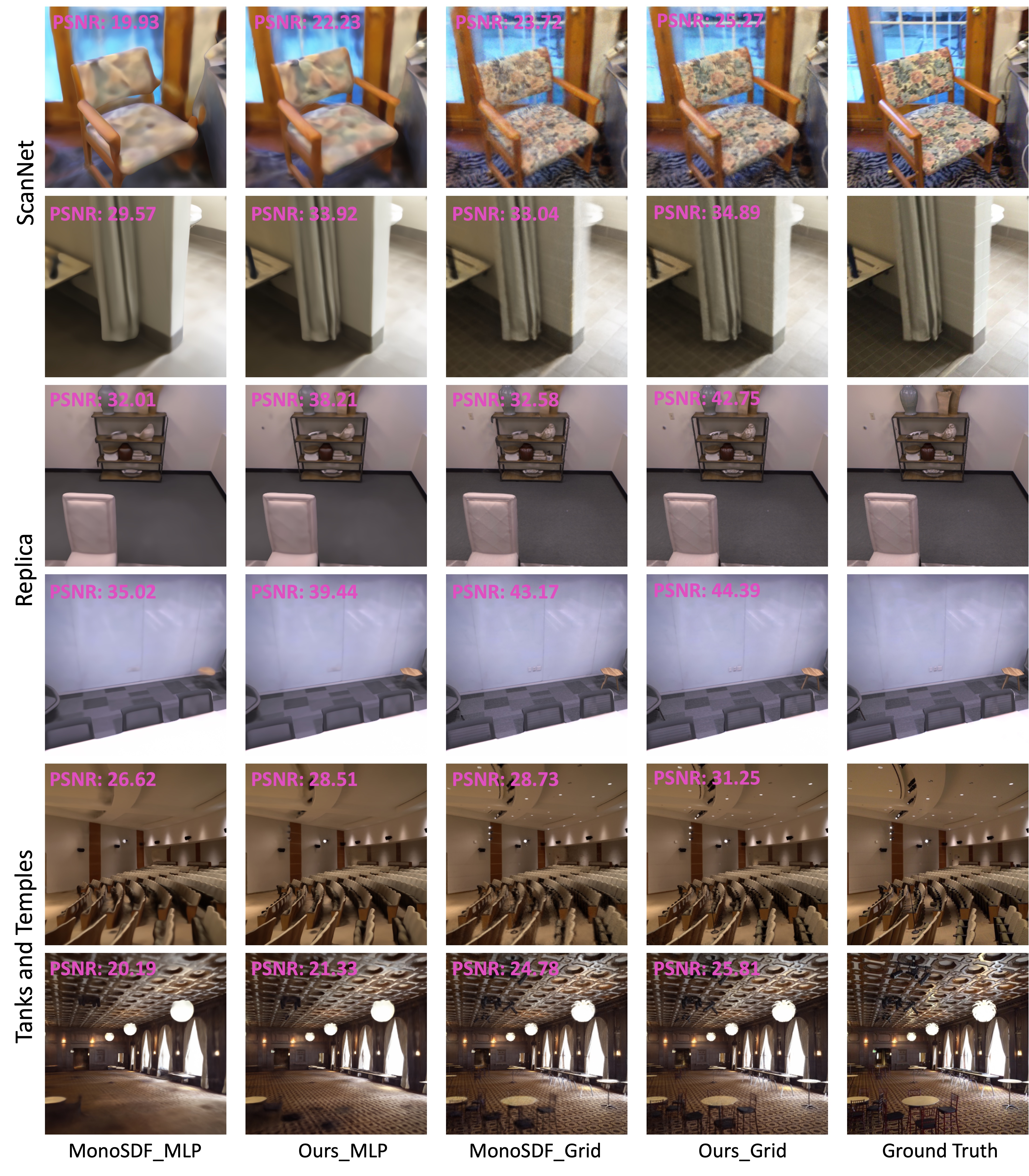}
  \caption{\textbf{Visualization of rendered views.} Our method improves the quality of generated views.
  }
  \label{fig:all_views}
\end{figure}

\cref{fig:all_views} presents visualization for rendered views. MonoSDF with MLP representation produces blurred regions, whereas our method captures more details and textures in the rendered views. Similarly, with the grid representation, our approach generates more precise views.

\section{Additional Analysis}
\label{sec:analysis}

\subsection{Analysis for Reconstruction}
The reconstruction task aims to predict accurate SDF values. Existing NeRF-based reconstruction methods \cite{yariv2021volume, guo2022neural} typically apply a 2D color loss as the primary loss function, in which the predicted color is rendered with volume density derived from SDF. However, the 2D color loss aims to learn the actual appearance, potentially interfering with the optimization of 3D geometry. For example, in the toy scene depicted in Fig. {1}, the 2D color loss guides the network to assign a higher weight to the point $\textbf{P}$ and a lower weight to the point $\textbf{Q}$. Consequently, the lower weight for the point $\textbf{Q}$ encourages the network to learn a SDF with a larger absolute value, contradicting the actual SDF for $\textbf{Q}$. This discrepancy negatively impacts surface reconstruction, particularly for small objects and thin regions.

Our method, including the ray-conditioned density function with SRDF, the SRDF-SDF consistency loss, and the self-supervised visibility task, can solve the aforementioned issue, generating more accurate surface geometry. The reasons are as follows: \textbf{(1) Reduced negative impact from 2D output:} Our approach utilizes SRDF to represent density, guiding the network to predict a positive SRDF with a large absolute value at point $\textbf{Q}$, consistent with SRDF's definition. This partially mitigates the negative influence of 2D losses on SDF. The comparison between \cref{fig:ab} - \emph{(Baseline)} and \cref{fig:ab} -\emph{(RS. Den.)} also proves that our method can generate better geometry, \eg recalling thin objects. \textbf{(2) Positive impact from the proposed SRDF-SDF consistency loss:} As analyzed in Fig. {1}, the density generated from the SRDF closely matches the actual observation. Hence, with multi-view input, the network can learn an accurate SRDF and density distribution among the entire 3D space. However, the network may struggle to produce precise SDF, resulting in missed surfaces, as shown in \cref{fig:ab} -\emph{(RS. Den.)}. Our SRDF-SDF consistency loss is designed to ensure SDF shares the same sign as SRDF, which can facilitate the recall of missed surfaces. As observed in \cref{fig:ab} -\emph{(RS. Den. + Con. L.)}, more surfaces are recalled. \textbf{(3) Positive impact from the proposed self-supervised visibility task:} Our self-supervised visibility task incorporates the physical visibility prior into the network, which can help the learning of SRDF and SDF. The comparisons in \cref{fig:ab} demonstrate that our self-supervised visibility task can recover more surfaces. Overall, our proposed method contributes to producing accurate SDF and reconstructing superior 3D geometry.

\subsection{Analysis for View Synthesis}
Existing methods \cite{yariv2021volume, guo2022neural} utilize SDF to generate the volume density. However, as depicted in Fig. {1} and Fig. {5}, the transformed density and weight derived from predicted SDF may introduce noises during volume rendering, resulting in inaccurate 2D color and consequently low PSNR. In contrast, our approach generates volume density from SRDF when rendering 2D color, ensuring consistent weights with actual observations and avoiding the influence of SDF. As a result, views rendered by our method achieve higher PSNR.

\end{document}